\documentclass[a4paper,fleqn]{cas-sc}

\usepackage[authoryear]{natbib} 

\usepackage{xcolor}
\usepackage{color}

\usepackage{algorithm}
\usepackage{algorithmic}

\usepackage{amssymb}
\usepackage{amsmath}
\usepackage{booktabs} 
\usepackage{longtable} 
\usepackage{multirow} 
\usepackage{array} 
\usepackage{graphicx}
\usepackage{subcaption}
\usepackage{float}
\usepackage{hyperref}
\usepackage{placeins}
\usepackage{comment}
\usepackage{rotating}
\usepackage{microtype}
\usepackage[capitalise]{cleveref}

\def\tsc#1{\csdef{#1}{\textsc{\lowercase{#1}}\xspace}}
\tsc{WGM}
\tsc{QE}

\begin{document}

\shorttitle{Navigational Safety of Marine Surface Vehicles}	

\shortauthors{} 

\title [mode = title]{Navigational Safety of Marine Surface Vehicles in Restricted Visibility: From Intelligent Vision to Vision-Centered Multi-Sensor Fusion}



%

\author[1,2,3,4]{Ryan Wen Liu}
\ead{wenliu@whut.edu.cn}

\author[2,4]{Fengchuan Song}
\ead{fc_song@whut.edu.cn}

\author[2,4]{Tianxiang Wang}
\ead{tianxiangwang@whut.edu.cn}

\author[2,4]{Guobei Peng}
\ead{guobeipeng@whut.edu.cn}

\author[5]{Youngrong Kim}
\ead{youngrong.kim@ntu.edu.sg}

\author[1,4,6]{Xian Zhong}
\ead{zhongx@whut.edu.cn}

\author[7]{Hooi-Siang Kang}
\ead{kanghs@utm.my}

\author[1,2,4]{Maohan Liang}
\cormark[1]
\ead{mhliang@whut.edu.cn}

\author[8]{Junzhi Yu}
\cormark[1]
\ead{yujunzhi@pku.edu.cn}

\author[2,9]{Shanliang Yao}
\cormark[1]
\ead{shanliang.yao@ycit.edu.cn}

\address[1]{Sanya Science and Education Innovation Park of Wuhan University of Technology, Sanya 572025, China}

\address[2]{School of Navigation, Wuhan University of Technology, Wuhan 430063, China}

\address[3]{School of Marine Science and Technology, Wuhan University of Technology, Sanya 572025, China}

\address[4]{State Key Laboratory of Maritime Technology and Safety, Wuhan 430063, China}

\address[5]{School of Civil and Environmental Engineering, Nanyang Technological University, 639798, Singapore}

\address[6]{Hubei Key Laboratory of Transportation Internet of Things, School of Artificial Intelligence, Wuhan University of Technology, Wuhan 430070, China}

\address[7]{Faculty of Mechanical Engineering, Universiti Teknologi Malaysia, Johor Bahru, Malaysia}

\address[8]{Department of Advanced Manufacturing and Robotics, College of Engineering, Peking University, Beijing 100871, China}

\address[9]{School of Information Engineering, Yancheng Institute of Technology, Yancheng 224051, China}

\cortext[1]{Corresponding authors}



\begin{abstract}
Marine surface vehicles (MSVs) have attracted considerable attention for their commercial, scientific, and military applications. Owing to their favorable performance-to-cost ratio, shipboard cameras are widely used to detect and recognize potentially hazardous situations across diverse operational scenarios. However, visual perception is inevitably degraded by restricted conditions, including irregular motion of the imaging platform, adverse weather, and long imaging distances. To improve situational awareness and navigation capabilities, complementary sensors, such as the automatic identification system (AIS), radar, and light detection and ranging (LiDAR), \textit{etc.}, have become essential shipboard equipment. Multi-sensor fusion can provide richer perceptual information and help ensure navigational safety of MSVs in restricted visibility. This article reviews recent progress from intelligent vision to vision-centered multi-sensor fusion for newcomers and researchers in intelligent ship navigation and intelligent waterborne transportation system (IWTS). After introducing the generation of quality-degraded images in restricted visibility, we will review intelligent vision methods for four key issues, \textit{i.e.}, shipboard video stabilization, visual perception enhancement, object detection and tracking, and navigable area segmentation. We then examine vision-centered multi-sensor fusion methods, including infrared-visible image fusion, radar-vision fusion, LiDAR-vision fusion, AIS-vision fusion, and other multi-sensor fusion scenarios. Open-source datasets and safety-critical applications are also summarized and discussed. Finally, we highlight open challenges and future research directions. The collected maritime datasets are available at \url{https://github.com/MAPGroup/Datasets} and will be regularly updated.
\end{abstract}




\maketitle

\section{Introduction}\label{sec:introduction}
%
%
%
\begin{figure}
	\centering
	\includegraphics[width=0.97\linewidth]{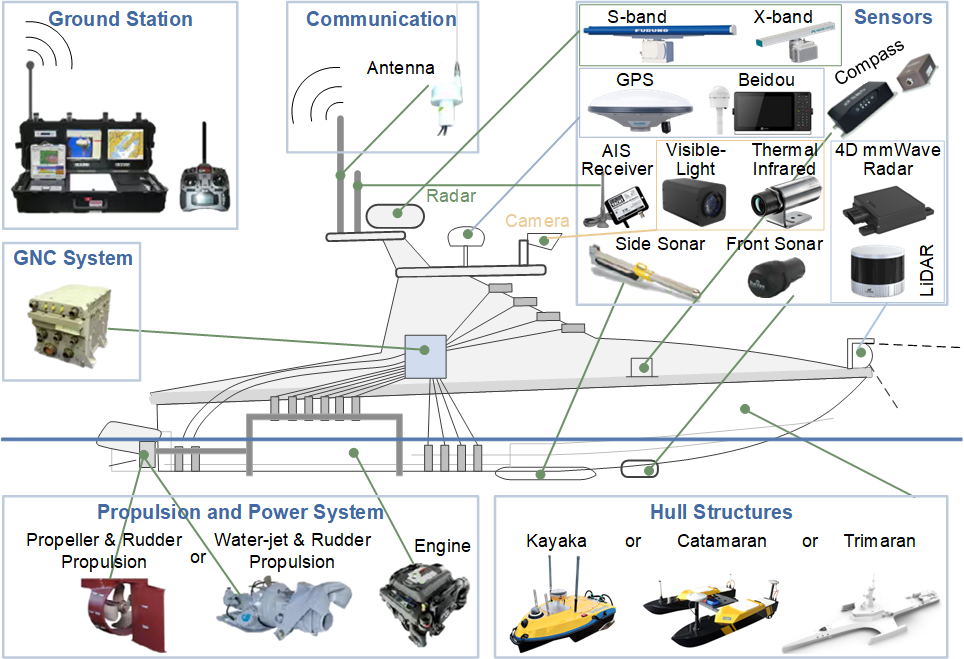}
	\caption{\textbf{Fundamental architecture of a typical MSV.} Visual illustration of an MSV equipped with multiple sensors, a ground station, GNC (guidance, navigation and control) system, communication system, propulsion and power system. Please refer to Ref.~\citep{liu2016unmanned} for more details on the typical elements of MSV, \textit{i.e.}, sensors (\textit{a.k.a.}, sensing data collection equipment), hull and auxiliary structural elements, propulsion and power system, GNC system, communication system, ground station, \textit{etc}.}
	\label{fig:Figure_MSV}
\end{figure}
\subsection{Background}
Marine surface vehicles (MSVs), including unmanned surface vehicles (USVs) and autonomous surface vehicles (ASVs), have attracted considerable attention from both academia and industry, as illustrated in \cref{fig:Figure_MSV}. Rapid advances in perception, communication, control, and artificial intelligence (AI) have enabled higher levels of safety, security, and efficiency in complex maritime environments~\citep{forti2022next}. Consequently, MSVs can perform dangerous, dull, and dirty tasks with minimal or no human intervention, supporting applications such as environmental monitoring, hydrologic surveys, search and rescue, military reconnaissance, and defense. However, the corresponding operational effectiveness will be severely affected by adverse weather conditions (\textit{e.g.}, haze, rain, snow, and low-lightness, \textit{etc}.), which significantly reduce the perception range, accuracy, and reliability for MSVs in practical applications. To guarantee reliable and safe navigation in restricted visibility, many efforts have focused on how to optimally select the shipborne sensors, which enable MSVs to increase the situational awareness capacity. In addition, how to develop the suitable and effective computing methods, related to shipborne sensors, is another long-standing challenge which urgently needs to be handled. The corresponding computing methods should exploit the full advantages of different sensors, and then provide high-level intelligence, high-precision perception, and high-reliability decision-making for MSVs in dynamic maritime conditions. 

\begin{sidewaystable}[!htbp]
\caption{\textbf{Comparisons of several shipboard sensors for MSVs.} The visible-light camera, thermal infrared camera, AIS (automatic identification system), S-band radar, X-band radar, 3D LiDAR (light detection and ranging), and 4D mmWave (millimeter wave) radar are compared and analyzed in terms of data dimension, spatial resolution, temporal resolution, detection range, cost, interference sources, strong points, and application scenarios. Note that the visible-light camera, thermal infrared camera, and AIS belong to passive sensors. In contrast, the S/X-band radar, 3D LiDAR, and 4D mmWave radar are active sensors, which perceive their environment through the actively emitted electromagnetic waves.}
\centering
\footnotesize
\setlength{\tabcolsep}{2pt}
\renewcommand{\arraystretch}{1.12}
\begin{tabular}{@{}>{\raggedright\arraybackslash}p{0.10\textheight}>{\raggedright\arraybackslash}p{0.095\textheight}>{\raggedright\arraybackslash}p{0.11\textheight}>{\raggedright\arraybackslash}p{0.055\textheight}>{\raggedright\arraybackslash}p{0.065\textheight}>{\raggedright\arraybackslash}p{0.05\textheight}>{\raggedright\arraybackslash}p{0.15\textheight}>{\raggedright\arraybackslash}p{0.12\textheight}>{\raggedright\arraybackslash}p{0.18\textheight}@{}}
\toprule[1.1pt]
\textbf{Sensors} & \textbf{Data}\newline\textbf{Dimension} & \textbf{Spatial}\newline\textbf{Resolution} & \textbf{Temporal}\newline\textbf{Resolution} & \textbf{Detection}\newline\textbf{Range} & \textbf{Cost} & \textbf{Interference}\newline\textbf{Sources} & \textbf{Strong}\newline\textbf{Points} & \textbf{Application}\newline\textbf{Scenarios} \\
\midrule
\textbf{Visible-Light}\newline\textbf{Camera} & 2D planar image & Pixel-level (1K/2K/4K) & High & Medium & Low--Medium & Rain/snow/haze/low-light attenuation, flare/glare, \textit{etc}. & High-resolution images & Daytime target detection, short-range channel monitoring, berthing/unberthing visual assistance, target identity verification \\
\addlinespace[2pt]
\textbf{Thermal Infrared}\newline\textbf{Camera} & 2D thermal radiation image & Pixel-level\newline{ (640$\times$512) }& High & Medium & Medium--High & High-temperature background interference, heavy rainfall attenuation, \textit{etc}. & Almost all-weather detection capability & Nighttime target detection, auxiliary observation in adverse weather, maritime search and rescue \\
\addlinespace[2pt]
\textbf{AIS} & 1D attribute data & $\sim$10 m \newline{(depends on GNSS)} & Low & Long & Low & VHF electromagnetic interference, AIS disabled by vessels, false signals, \textit{etc}. & Collision avoidance and largely unaffected by adverse weather & Vessel identity recognition, situational awareness, collision avoidance, \textit{etc}. \\
\addlinespace[4pt]
\textbf{S-band Radar}\newline\textbf{(2--4 GHz)} & 2D planar plots/tracks \newline{(range, azimuth)} & Range resolution 10--30 m; azimuth resolution 1$^\circ$--2.3$^\circ$ & High & Long & Medium--High & Sea clutter, rain attenuation, same-frequency interference, \textit{etc}. & Robust long-range target detection in adverse weather & Medium- to long-range wide-area detection, ocean-going collision avoidance, maritime surveillance, \textit{etc}. \\
\addlinespace[2pt]
\textbf{X-band Radar}\newline\textbf{(8--12 GHz)} & 2D planar plots/tracks \newline{(range, azimuth)} & Range resolution 1--15 m; azimuth resolution 0.5$^\circ$--1$^\circ$ & High & Long & Medium & Sea clutter, rain/snow attenuation, same-frequency interference, \textit{etc}. & Smaller antenna and higher target resolution & Short- to medium-range high-precision detection, small-target identification, collision avoidance in narrow channels, \textit{etc}. \\
\addlinespace[2pt]
\textbf{3D LiDAR} & 3D point cloud (range, azimuth, elevation, reflection intensity) & Range resolution $\sim$2 cm; angular resolution 0.01$^\circ$--0.05$^\circ$ & High & Short--Medium & High & Rain/snow/haze attenuation, strong light interference, water surface reflection, \textit{etc}. & Highly accurate surface measurements and 3D mapping & Short-range high-precision 3D modeling, berthing/unberthing assistance, \textit{etc}. \\
\addlinespace[2pt]
\textbf{4D mmWave}\newline\textbf{Radar} & 4D point cloud (range, azimuth, elevation, radial velocity) & Range resolution 0.1--2 m; azimuth resolution 1$^\circ$--3$^\circ$ & High & Short & Medium--High & Same-band millimeter-wave interference, heavy rainfall attenuation, \textit{etc}. & Almost all-weather operation and robust perception & Short- to medium-range multi-target detection, collision avoidance in adverse weather, \textit{etc}. \\
\bottomrule[1.1pt]
\end{tabular}
\label{tab:SensorsComparison}
\end{sidewaystable}

In unknown and complex environments, the operational capabilities of MSVs, including situational awareness, decision-making, and control, depend heavily on shipborne sensors, such as visible-light and thermal infrared cameras, automatic identification system (AIS), radar, and light detection and ranging (LiDAR), \textit{etc}. Visible-light camera, which operates similarly to human vision, provides detailed red-green-blue (RGB) images. However, the imaging quality will be significantly degraded under low-visibility weathers. In contrast, thermal infrared camera could capture the shapes of objects of interest through their emitted infrared radiation. Owing to the capacity of nighttime target detection, it has played a significant role in enhancing the navigational safety of MSVs. AIS, an automated vessel self-reporting system, receives both static information (\textit{e.g.}, the maritime mobile service identity (MMSI), vessel name, length, and breadth) and dynamic information (\textit{e.g.}, position, course, and speed) from surrounding vessels through very-high-frequency (VHF) radio signals~\citep{liu2020data}. It has been widely applied to traffic-risk analysis, collision avoidance, and route planning~\citep{liu2022ship}. Radar detects surrounding surface vehicles and provides their bearings and distances for collision avoidance and navigation. In particular, S-band radar is able to robustly detect targets during poor weathers (\textit{e.g.}, hazy, snow and rain, \textit{etc}.) since it has a longer wavelength. However, it offers lower resolution for detecting small targets. X-band radar performs well in detecting small targets (\textit{e.g.}, small ships and buoys, \textit{etc}.) but will fail in poor visibility conditions. In contrast, LiDAR uses laser rather than radio waves for short-range three-dimensional environmental perception, resulting in short-range high-precision surface measurements and 3D mapping capabilities. 4D millimeter wave (4D mmWave) radar has the capacity of measuring the range, azimuth, elevation, and velocity of targets \citep{yao2024waterscenes}. It has attracted considerable attention in autonomous driving because of its all-weather adaptability and velocity-measuring capability. More details on the comparison of different shipborne sensors could be found in Table \ref{tab:SensorsComparison}.

Owing to the favorable performance-to-cost ratio and installation convenience, visible-light camera has been widely exploited to support safe navigation. However, its imaging quality is readily degraded by adverse weather and illumination conditions, including low illumination, haze, snow, and rain~\citep{cao2026visual}. Accordingly, visual perception enhancement methods have been extensively studied to improve navigational safety of MSVs in restricted visibility. With the rapid development of artificial intelligence (AI) and computer vision (CV) techniques, many advanced Visual computing methods, including maritime object detection, recognition, tracking, and segmentation, further help MSVs understand their surroundings in diverse imaging and weather conditions. More recently, multi-sensor data fusion has received increasing attention for exploiting complementary sensing capabilities. Because cameras provide rich color and semantic information, this survey mainly focuses on vision-centered multi-sensor data fusion, including infrared-visible image fusion, radar-vision fusion, LiDAR-vision fusion, and AIS-vision fusion, \textit{etc}.%
\subsection{Motivation and Contributions}
%
%
By jointly exploiting heterogeneous sensor data to form a more reliable representation of the navigational environment, multi-sensor data fusion has become a key enabling technology for reliable assisted and autonomous navigation. Substantial efforts have surveyed data fusion for autonomous driving, intelligent vehicles, and intelligent transportation systems. Existing studies have demonstrated that fusion strategies can improve perceptual accuracy and decision reliability, thereby enhancing system reliability, robustness, and safety. To the best of our knowledge, no systematic survey has jointly reviewed intelligent vision and vision-centered data fusion for ensuring the navigational safety of MSVs in complex maritime environments.

\begin{figure}
	\centering
	\includegraphics[width=\linewidth]{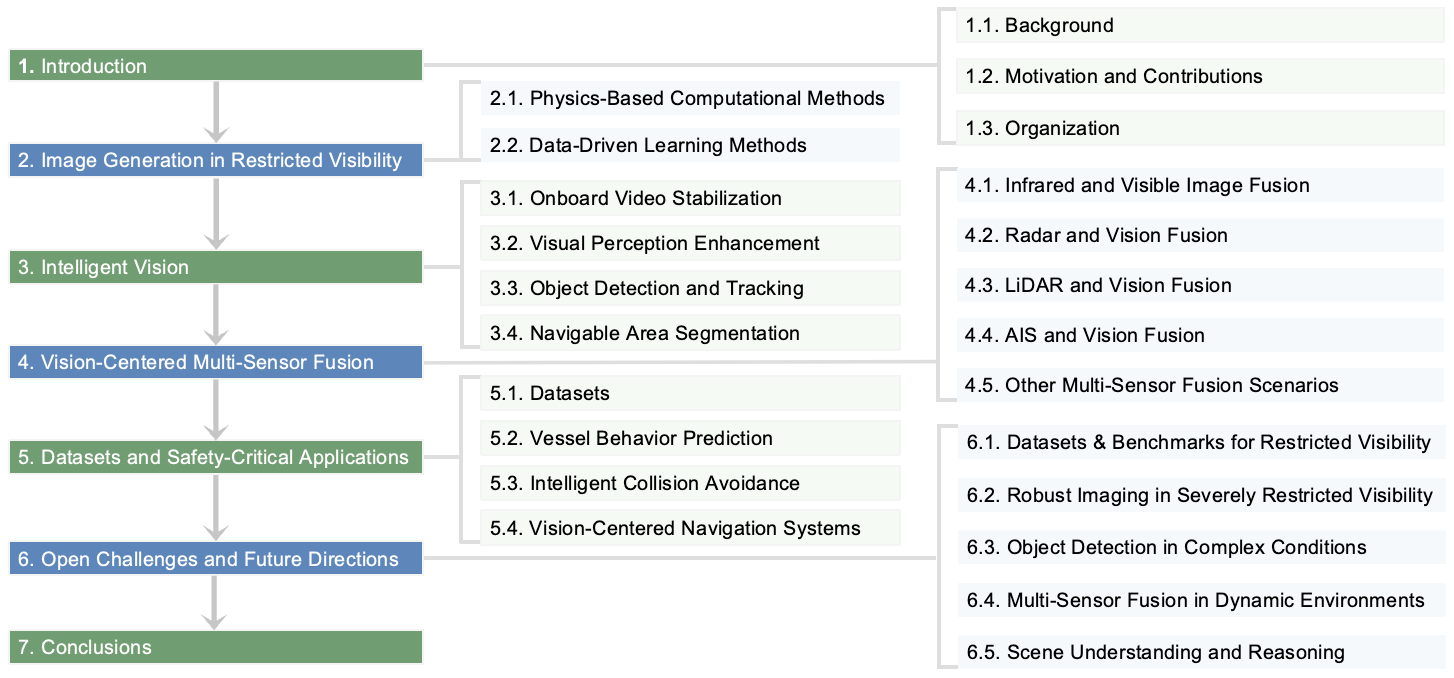}
	\caption{\textbf{Organizational structure of our work.} The image generation, intelligent vision, and vision-centered multi-sensor fusion methods, as well as the maritime datasets and safety-critical practical applications are firstly discussed. We then analyze the open challenges and future research directions in this work.}
	\label{fig:StructureOurWork}
\end{figure}

Compared with road traffic, maritime traffic is more complex and vulnerable because of adverse weather and illumination conditions, high vessel-traffic density, and the unstable waterborne operating environment. Despite rapid advances in intelligent shipping technologies, maritime accidents can cause environmental pollution, property damage, injury, and loss of life. To improve navigational safety, numerous AI-based computational methods have been developed to enhance intelligent visual perception. We therefore first review representative intelligent vision methods, including shipboard video stabilization, visual perception enhancement, object detection and tracking, and navigable area segmentation, shown in Fig. \ref{fig:StructureOurWork}. These methods enhance visual perception in restricted visibility, particularly in narrow waterways, ports, and congested waters. We then survey vision-centered multi-sensor data fusion methods, including infrared-visible image fusion, radar-vision fusion, LiDAR-vision fusion, and AIS-vision fusion. Their reliability and applicability have been evaluated through real-world experiments reported in the literature.

The main contributions of this work are summarized as follows:
\begin{itemize}
    \item We present a structured survey that connects intelligent vision with vision-centered multi-sensor data fusion, providing a unified perspective on perception techniques for safe MSV navigation in complex maritime and restricted-visibility conditions.
   \item We systematically review maritime perception and multi-sensor fusion datasets and synthesize applications in behavior prediction, collision avoidance, and autonomous navigation, highlighting the connection between environmental perception and safety-critical navigation functions.
    \item We identify key open challenges and future research directions, including restricted-visibility benchmarks, robust imaging under severe degradation, small-object detection, uncertainty-aware multi-sensor fusion, and vision-language scene understanding.
\end{itemize}

This work is intended to provide a comprehensive reference for newcomers and researchers studying the navigational safety of unmanned and autonomous surface vehicles in restricted visibility.%
%
%

\begin{table}[!t]
\caption{\textbf{Abbreviations used in this article.}}
\centering
\setlength{\tabcolsep}{12pt}
\footnotesize
\begin{tabular}{l|l|l|l}
\toprule[1.1pt]
Abbreviation & Full term & Abbreviation & Full term \\
\midrule
AI   & Artificial intelligence
     & CV   & Computer vision \\
AIS  & Automatic identification system
     & MMSI & Maritime mobile service identity \\
ASVs & Autonomous surface vehicles
     & USVs & Unmanned surface vehicles \\
MSVs & Marine surface vehicles
     & MASS & Maritime autonomous surface ship \\
IMU  & Inertial measurement unit
     & VHF  & Very-high-frequency \\   
UHD  & Ultra-high definition
     & mmWave & Millimeter-wave \\ 
FOV  & Field of View
     & DOF  & Degree-of-freedom \\
BEV  & Bird’s-eye view 
     & ROS  & Robot operating system \\
LiDAR& Light detection and ranging
     & AR   & Augmented-reality \\
GCN  & Graph convolutional network      
     & GAN  & Generative adversarial network \\   
HADAR& Heat-assisted detection and ranging     
     & MLLM & Multimodal large language model \\ 
DCPA & Distance at the closest point of approach     
     & TCPA & Time to closest point of approach \\ 
IWTS & Intelligent waterborne transportation system
     & mIoT & Maritime Internet of Things \\
\bottomrule[1.1pt]
\end{tabular}
\label{tab:abbreviations}
\end{table}

\subsection{Organization}
The remainder of this paper is organized as follows. \cref{sec:preliminaries} introduces physical modeling and data-driven methods for generating visually degraded images in restricted visibility. \cref{sec:intelvision} reviews recent studies on intelligent vision, and \cref{sec:datafusion} analyzes vision-centered multi-sensor data fusion methods. \cref{sec:datasets} discusses the multi-modal datasets and safety-critical maritime applications. We further discuss the open challenges and future research directions in \cref{sec:challenging}. \cref{sec:conclusions} concludes the paper by summarizing its main contributions. \cref{tab:abbreviations} lists the principal abbreviations and acronyms used throughout this paper.

%
\section{Image Generation in Restricted Visibility} \label{sec:preliminaries}
\begin{figure}
	\centering
	\includegraphics[width=\linewidth]{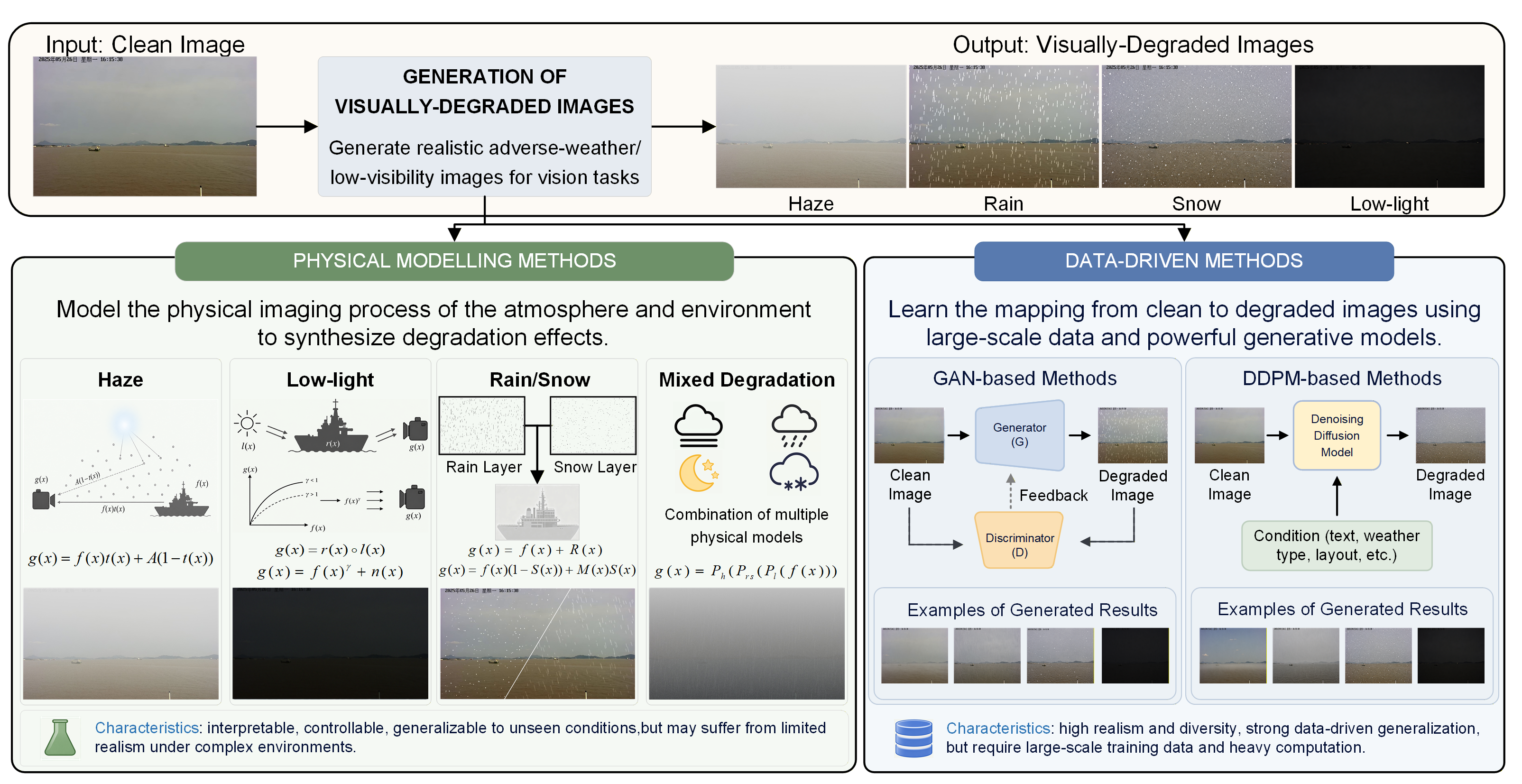}
	\caption{\textbf{Overview of visually degraded image generation methods.} Existing methods are mainly divided into two categories, \textit{i.e.}, physical modeling and data-driven methods.}
	\label{fig:Figure_03}
\end{figure}

Before reviewing intelligent vision and vision-centered multi-sensor data fusion, we introduce physics-based and data-driven methods for generating visually degraded images in restricted visibility. These methods can increase the volume and diversity of training data, thereby improving the robustness and generalization of intelligent vision and data fusion methods. As illustrated in \cref{fig:Figure_03}, existing approaches can be broadly divided into physics-based and data-driven methods. Physics-based methods synthesize degradations according to imaging mechanisms, whereas data-driven methods learn clean-to-degraded mappings from large-scale data using generative models.

\subsection{Physics-Based Computational Methods}
Visually degraded images can be synthesized using the following physical imaging models.
\subsubsection{Hazy Imaging Model}

According to Koschmieder's law~\citep{galdran2015enhanced}, a synthetic hazy image $g$ can be formulated as
\begin{align} \label{eq:hazymodel} 
g \left(\mathrm{x} \right) = f \left(\mathrm{x} \right)t \left(\mathrm{x} \right) + A \left(1-t \left(\mathrm{x} \right) \right), 
\end{align} 
where $\mathrm{x}\in\Omega$, with $\Omega$ denoting the image domain; $f$ is the latent haze-free image; and $A\in(0,1]$ and $t\in(0,1]$ denote the atmospheric light and transmission map, respectively. A lower transmission indicates more severe visual degradation, which can impair subsequent detection and tracking.
\subsubsection{Low-Light Imaging Model}
According to Retinex theory~\citep{land1977retinex}, a synthetic low-light image $g$ can be modeled as
\begin{align}\label{eq:lowlightRetinexmodel}
g \left(\mathrm{x} \right) = r \left(\mathrm{x} \right) \circ l \left(\mathrm{x} \right),
\end{align}
where $\circ$ denotes pixel-wise multiplication, and $r$ and $l$ are the reflectance and illumination components, respectively. During training, images can be randomly transformed into low-light versions before each epoch using \cref{eq:lowlightRetinexmodel} with $l\in(0,1)$.

Gamma-correction-based degradation is another widely-used strategy for synthesizing low-light images, \textit{i.e.},
\begin{align}\label{eq:lowlightGammamodel}
g \left(\mathrm{x} \right) = f \left(\mathrm{x} \right)^\gamma + n \left(\mathrm{x} \right),
\end{align}
where $\gamma$ is the gamma-correction coefficient and $n$ denotes sensor noise.
\subsubsection{Rainy or Snowy Imaging Model}
Rain and snow degradations are commonly synthesized by adding weather-related degradation layers. A rainy image $g$ can be formulated as
\begin{align}\label{eq:rainmodel}
g \left(\mathrm{x} \right) = f \left(\mathrm{x} \right) + \mathcal{R} \left(\mathrm{x} \right),
\end{align}
where $\mathcal{R}$ denotes the rain-streak layer. This model assumes that rain streaks primarily introduce high-frequency texture interference through linear superposition.

Because snow particles produce stronger occlusion and translucency effects, a snowy image $g$ can be modeled as
\begin{align}\label{eq:snowmodel}
g \left(\mathrm{x} \right) = f \left(\mathrm{x} \right) \left(1-\mathcal{S} \left(\mathrm{x} \right) \right) + \mathit{M} \left(\mathrm{x} \right)\mathcal{S} \left(\mathrm{x} \right), 
\end{align}
where $\mathcal{S}$ denotes the snow mask and $\mathit{M}$ represents the snow chromatic mapping component.
\subsubsection{Mixed Degradation Model}
In real-world environments, visual degradation is often caused by coupled adverse factors, including haze scattering, low illumination, rain, and snow. Such complex conditions can be synthesized by cascading the corresponding elementary degradation operators into a unified physical model, \textit{i.e.},
\begin{align}\label{eq:{mixeddegradationmodel}}
g \left(\mathrm{x} \right) = \mathcal{P}_h \left(\mathcal{P}_{rs} \left(\mathcal{P}_l \left(f \left(\mathrm{x} \right) \right) \right) \right),
\end{align}
where $\mathcal{P}_h$, $\mathcal{P}_{rs}$, and $\mathcal{P}_l$ denote the haze, rain-or-snow, and low-light degradation operators, respectively. Cascading different degradation operators enables this formulation to represent complex adverse-weather imaging conditions more effectively.
\subsection{Data-Driven Learning Methods}
Unlike physics-based approaches, data-driven methods use large-scale datasets to generate visually realistic degradations without explicitly modeling the physical imaging process. Although most existing studies focus on road transportation, these weather-generation techniques can also inform maritime vision tasks because of their similar atmospheric degradation characteristics. Generative adversarial networks (GANs) and denoising diffusion probabilistic models (DDPMs) are two mainstream paradigms for adverse-weather scene generation.
\subsubsection{GAN-Based Methods}
GANs learn the distribution of real images through adversarial training between a generator and a discriminator, enabling visually realistic image synthesis and providing a methodological foundation for degraded image generation~\citep{Goodfellow2014generative}. GAN-based methods can be broadly divided into paired clear-to-degraded image translation and unpaired domain translation.

Paired clear-to-degraded translation learns mappings using paired source and target images. A representative method is pix2pix, which learns a source-to-target domain mapping through paired supervision~\citep{isola2017image}. For degraded image generation, clear images are used as inputs and images of the same scenes in restricted visibility serve as targets. This paradigm supports data augmentation and annotation reuse while largely preserving scene structures. For example, \citet{shao2022high} developed a multi-scale pix2pix-based model to learn mappings between clear and rainy driving images, thereby generating high-quality rainy scenes. \citet{li2024gan} used the pix2pix framework for controllable augmentation of low-light roadside images, synthesizing nighttime or low-visibility traffic scenes to improve object detection. Paired translation provides explicit mappings and strong structure preservation, but its practical use is constrained by the difficulty of collecting strictly paired clear and degraded images.

Unpaired domain translation learns mappings without paired samples. CycleGAN, a representative framework, uses adversarial and cycle-consistency constraints to learn mappings between source and target domains~\citep{zhu2017unpaired}. Because independently collecting clear and degraded images is generally easier than obtaining strict pairs, this paradigm is more suitable for real restricted-visibility scenarios. Weather GAN supports translation among sunny, cloudy, foggy, rainy, and snowy domains and reduces structural distortion in non-degraded regions through weather-cue segmentation and attention modules~\citep{li2021weather}. SUSTechGAN targets rainy and nighttime image generation for autonomous driving by incorporating dual-attention modules, a multi-scale generator, and task-related losses; the generated images are further used to improve object detection in adverse conditions~\citep{lan2024sustechgan}. Although unpaired methods avoid reliance on strictly paired data, they may provide imprecise control over degradation intensity and cause object-boundary deformation or semantic-structure drift.
\subsubsection{DDPM-Based Methods}
Diffusion models~\citep{ho2020denoising} have been widely applied to visual generation because of their training stability and generation quality. With advances in vision-language representations, text conditioning has been incorporated into diffusion models to control image content, scene attributes, and degradation conditions, such as weather, noise, and blur~\citep{ramesh2022hierarchical,rombach2022high}. This provides a flexible basis for semantically controlled degraded image generation. According to whether additional non-textual conditions, such as images, degradation intensity, and layouts, are incorporated, these methods can be divided into text-only conditioned and multi-conditioned approaches.

Text-only conditioned methods use manually specified prompts or prompts generated by large language models to control degradation types and scene attributes. For example, \citet{rothmeier2024time} propose a text-to-image model for synthesizing driving scenes under snow, haze, and rain. The generated samples are subsequently used to fine-tune object detectors, consistently improving detection accuracy across adverse weather conditions. WeatherDG combines Stable Diffusion with a large-language-model-based procedural prompt-generation strategy to synthesize driving scenes under snow, rain, haze, and low-light conditions, thereby improving the adverse-weather domain generalization of semantic segmentation models~\citep{qian2025weatherdg}. Text-only conditioning improves semantic control over degradation types and scene attributes, but may still produce inaccurate degradation regions, structural drift, and images unsuitable for annotation reuse.

Beyond text conditioning, multi-conditioned methods incorporate image content, degradation attributes, or scene structures to provide finer control over degradation and scene generation. GenDeg introduces a conditional diffusion model that accounts for both degradation type and intensity and can synthesize large-scale images covering six degradation types~\citep{rajagopalan2025gendeg}. GenDS, constructed using GenDeg, further demonstrates that all-in-one restoration models trained on the generated dataset achieve improved generalization. \citet{guo2025neptune} present X-to-Maritime, a multimodally conditioned generative model for maritime object detection that integrates text, object layouts, water-surface environments, and imaging attributes to generate diverse maritime scenes and improve detection generalization. Compared with text-only conditioned methods, multi-conditioned methods provide stronger control over degradation attributes, scene structures, and task-oriented requirements, but incur higher computational complexity because of the integration of multiple conditional inputs.
\section{Intelligent Vision}\label{sec:intelvision}

\begin{figure}
	\centering
	\includegraphics[width=\linewidth]{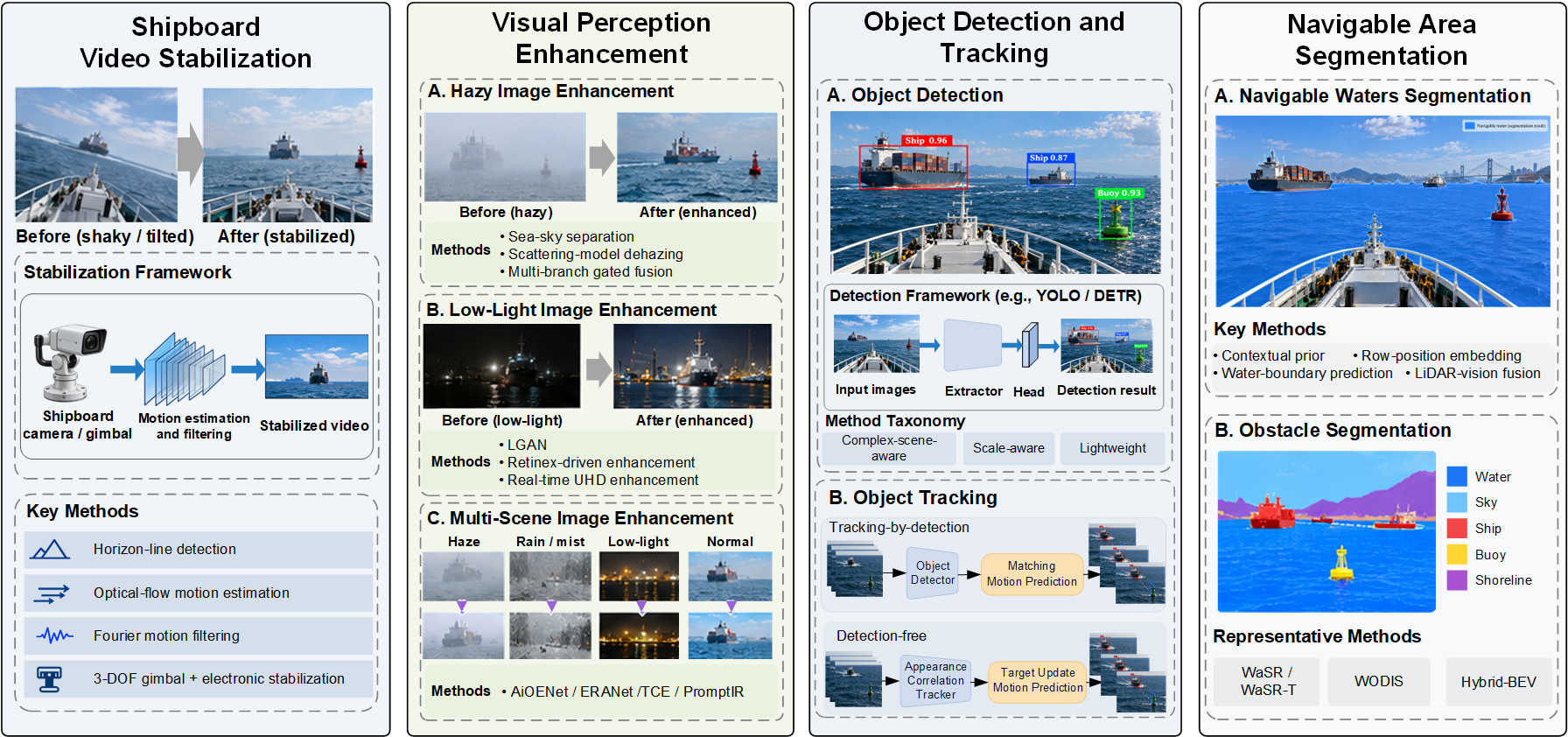}
	\caption{\textbf{Overview of intelligent vision for marine surface vehicles.} The intelligent vision pipeline integrates shipboard video stabilization, visual perception enhancement (including hazy, low-light, and multi-scene enhancement), object detection and tracking, and navigable area segmentation, providing robust situational awareness for MSVs in restricted visibility.}
	\label{fig:intelligent-vision}
\end{figure}

Adverse weather and illumination conditions can substantially degrade the visual perception of MSVs, increasing collision risks in complex and high-density waters. Integrating advanced intelligent vision modules into assisted and autonomous navigation systems is therefore important for robust situational awareness, as shown in \cref{fig:intelligent-vision}. This section reviews recent advances in shipboard video stabilization, visual perception enhancement, object detection and tracking, and navigable area segmentation.
%
%
\subsection{Shipboard Video Stabilization}
Video stabilization has been extensively studied across intelligent transportation systems. Traditional approaches can generally be divided into mechanical, optical, and electronic stabilization. For shipboard optical systems, many existing methods rely on accurate detection of the sea-sky horizon~\citep{zardoua2020short}. To facilitate experimentation and autonomous maritime navigation, the first public annotation software for horizon-line detection was developed and released in~\citep{zardoua2023horizon}. Schwendeman and Thomson proposed a method for reconfiguring and repairing shipboard videos by automatically tracking the dynamic horizon~\citep{schwendeman2015horizon}, together with an open-source Horizon Stabilization package covering horizon detection, image stabilization, rectification, and error analysis\footnote{The source code is available at \url{https://www.mathworks.com/matlabcentral/fileexchange/47606-horizon-stabilization}.}. \citet{wen2019feature} combined optical-flow-based motion estimation, Fourier-series-based motion filtering, and motion compensation to suppress random camera shake. To achieve real-time stabilization in heavy-sea environments, \citet{xin2022hierarchical} combined mechanical and electronic stabilization: a three-degree-of-freedom (DOF) gimbal with a robust controller provides primary motion compensation, followed by sea-sky-line-based electronic motion estimation and compensation. More advanced techniques developed for other stabilization tasks, including optical-flow-driven neural networks~\citep{yu2020learning}, meta-learning~\citep{ali2024harnessing}, GANs~\citep{xu2018deep}, and Transformers~\citep{wang2024soft}, may also be extended to shipboard video stabilization. Related limitations and emerging directions have been discussed in recent surveys~\citep{guilluy2021video,roberto2022survey,wang2023video}.
\subsection{Visual Perception Enhancement}
Shipboard visible-light cameras are frequently affected by adverse weather and imaging conditions, including haze, low illumination, rain, and snow. The resulting image degradation can impair downstream tasks such as object detection, recognition, tracking, and multi-sensor data fusion. This subsection reviews representative enhancement techniques for improving visual perception in these conditions.
%
%
\subsubsection{Hazy Image Enhancement}
Haze reduces the accuracy and robustness of maritime object detection and tracking and consequently weakens MSV situational awareness. Maritime scenes differ substantially from conventional natural scenes, and dehazing methods developed for non-maritime imagery may suffer from transmission-estimation errors or atmospheric-light misestimation. 
Because shipboard surveillance imagery is dominated by visually similar sea and sky regions, accurate sea-sky separation is also important. In~\citep{hu2023single}, mean-shift segmentation and adaptive threshold constraints were used to estimate the sea-sky separation and transmission map. Linear transformation and improved high-boost filtering were combined for maritime dehazing in~\citep{ma2016effective},
while a comprehensive scattering model integrating illumination decomposition, transmission estimation, and illumination compensation was further developed for haze removal under glow-shaped environmental illumination~\citep{hu2019single}. 

Deep learning has also been increasingly adopted for maritime image dehazing. \citet{fan2022multi} proposed an end-to-end multi-branch gated fusion network to handle high-frequency details, thick-haze regions, and contrast enhancement. \citet{guo2023haze} developed a dual-attention and dual-frequency-guided network to preserve both structural and low-frequency information while supporting real-time dehazing. To address non-uniform haze, \citet{liu2022deep} introduced a three-stage network that jointly estimates haze distribution, learns haze-related features, and performs multi-feature fusion. To alleviate the lack of paired hazy and clean ground truths, \citet{he2023single} developed an end-to-end GAN under unpaired supervision and constructed the realistic unpaired REMIDE dataset for maritime image dehazing.
\subsubsection{Low-Light Image Enhancement}
In low-light conditions, visible-light cameras receive limited illumination, resulting in low contrast and poorly resolved details that degrade both visual information and safety-related vision tasks. Low-light image enhancement aims to recover hidden information while suppressing unwanted noise and artifacts~\citep{li2021low}. It has been widely studied for intelligent ground vehicles~\citep{lu2025polarization}, unmanned aerial vehicles (UAVs)~\citep{wang2020low,xing2023clegan}, and autonomous underwater vehicles (AUVs)~\citep{im2024artificial}.

In maritime environments, high-value vessels may additionally employ cooled or uncooled thermal cameras. However, uncooled thermal cameras provide limited geometric texture, whereas cooled thermal cameras are considerably more expensive than conventional visible-light and uncooled thermal cameras. Owing to their low cost and convenient deployment, visible-light cameras therefore remain widely exploited for MSVs. Deep learning has shown strong performance for low-light image enhancement~\citep{liu2021benchmarking}. For maritime imagery, \citet{liu2022attention} proposed a lightweight GAN (LGAN) with an attention mechanism to suppress overexposure. To support object detection under low illumination, \citet{guo2022lightweight} developed a Retinex-driven lightweight network using depthwise separable convolution for low-light image enhancement. A robust variational regularized low-light enhancement method was presented by simultaneously refining the illumination and reflection maps~\citep{liu2024low}. In addition, a complexity-aware expert mechanism was developed to mitigate the negative effects of low-light images on maritime safety~\citep{he2025complexity}. As the spatial resolution of captured images increases (\textit{e.g.}, from standard definition of $480$p to ultra-high definition (UHD) of $4$K), the computational efficiency will be obviously decreased, leading to non-real-time image enhancement and unfriendly human-computer interaction. To obtain competitive enhancement quality and computational efficiency, a double domain guided real-time image enhancement network (DDNet) was proposed to handle UHD images under low-light conditions~\citep{qu2024double}.
\begin{figure}
	\centering
	\includegraphics[width=0.78\linewidth]{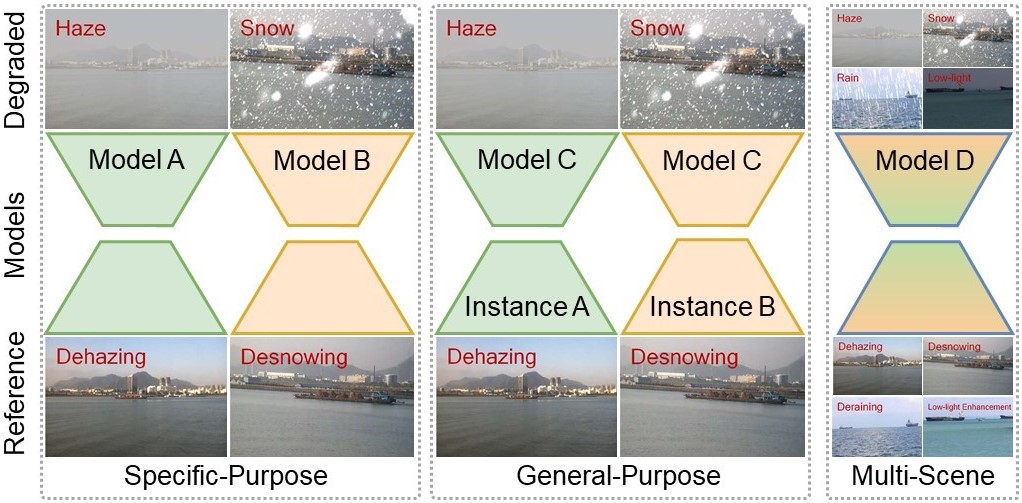}
	\caption{\textbf{Illustrations of \textit{specific-purpose}, \textit{general-purpose}, and \textit{multi-scene} image enhancement}. Each \textit{specific-purpose} model is tailored to handle a particular type of degradation (\textit{e.g.}, haze, snow, rain, and low-lightness). The \textit{general-purpose} model is proposed to tackle a variety of image enhancement tasks using a shared or scalable network~\citep{cui2026bio}. In contrast, the multi-scene framework is capable of tackling various degradations using one single model (\textit{a.k.a.}, all-in-one).}
	\label{fig:Figure_MultiSceneImage}
\end{figure}
\subsubsection{Multi-Scene Image Enhancement}
Most enhancement methods discussed above are \textit{specific-purpose}, targeting a single degradation type. To address multiple adverse conditions within a unified framework, \textit{multi-scene} enhancement framework has attracted increasing attention. \citet{liu2023aioenet} proposed an all-in-one low-visibility enhancement network (AiOENet) for haze, rain, snow, and low-lightness, although its large parameter count limits computational efficiency. To reduce computational cost for MSVs deployment in IWTS, \citet{liu2024real} developed an edge-reparameterization- and attention-guided network (ERANet), shown in \cref{fig:Figure_MultiSceneImage}, improving visual quality while maintaining low computational cost and supporting downstream object detection and scene segmentation under haze, rain, and low-light conditions. To promote vessel traffic safety in both hazy and low-light conditions, a dual-branch encoder–decoder architecture (TCE) was proposed to decompose the objective of \textit{multi-scene} enhancement into two synergistic sub-tasks, \textit{i.e.}, texture reconstruction and color restoration~\citep{zhang2026low}. The optimized results of these two sub-tasks jointly contribute to the final enhancement scenario. Other all-in-one restoration approaches, including PromptIR~\citep{potlapalli2024promptir}, ADMS~\citep{park2023all}, and AIRFormer~\citep{gao2023frequency}, have been developed for different application scenarios. Pre-trained vision-language models have also demonstrated the ability to restore images affected by multiple adverse-weather degradations~\citep{yang2024language}. These developments provide promising techniques for improving shipboard imaging quality across diverse degradation types and levels.
\subsection{Object Detection and Tracking}
Shipboard object detection and tracking face several recurring challenges, including substantial environmental variations, large target-scale differences, limited shipboard computational resources, and frequent occlusions. Recent advances in object detection and tracking are reviewed separately below.
\begin{table}[!t]
\caption{\textbf{Representative maritime object detection methods.}}
\centering
\setlength{\tabcolsep}{12pt}
\footnotesize
\begin{tabular}{lc|cl}
\toprule[1.1pt]
Method & Year & Modality & Framework \\
\midrule

\multicolumn{4}{l}{\textit{Complex-scene-aware methods}} \\

RetinaMFANet~\citep{ma2022improved}
& 2022 & VIS & RetinaNet \\

Weather-OD~\citep{chen2024weather}
& 2024 & VIS & Multiple detectors \\

MDD-ShipNet~\citep{wang2024mdd}
& 2024 & VIS & CNN \\

FREGNet~\citep{tian2024fregnet}
& 2024 & VIS & CNN-GCN \\

CFDBNet~\citep{tian2025coarse}
& 2025 & VIS & Dual-branch CNN \\

TSPNet~\citep{zhang2024tspnet}
& 2024 & IR video & Temporal pyramid \\

IS-MLDA~\citep{feng2024meta}
& 2024 & IR & YOLO-based \\

FEAFNet~\citep{zhang2024infrared}
& 2024 & IR & CNN-based \\

YOLO-IRS~\citep{guo2024yolo}
& 2024 & IR & YOLOv10 \\

CFLFE~\citep{ma2026spatio}
& 2026 & IR & Cascade Swin \\

\midrule
\multicolumn{4}{l}{\textit{Scale-aware methods}} \\

GAN-CNN detector~\citep{chen2020deep}
& 2020 & VIS & GAN-CNN \\

VarifocalNet~\citep{shao2022multi}
& 2022 & VIS & CNN-based \\

Multi-scale detector~\citep{zhang2023lightweight}
& 2023 & VIS & YOLOv7 \\

THFE~\citep{guo2023thfe}
& 2023 & VIS & CNN-based \\

YOLO-MRS~\citep{yu2024yolo}
& 2024 & VIS & YOLOv8 \\

RTDETR-MARD~\citep{sun2025rtdetr}
& 2025 & VIS & RT-DETR \\

USVRT-DETR~\citep{zhang2025usvrt}
& 2025 & VIS & RT-DETR \\

SARFB~\citep{wu2023sarfb}
& 2023 & IR & YOLOv5 \\

GLAF-DETR~\citep{zhang2025glaf}
& 2025 & IR & RT-DETR \\

IRSD-Net~\citep{sun2025irsd}
& 2025 & IR & YOLOv11 \\

FGOM-RTDETR~\citep{wang2026fgom}
& 2026 & IR & RT-DETR \\

\midrule
\multicolumn{4}{l}{\textit{Lightweight detection methods}} \\

Saliency-aware CNN~\citep{shao2019saliency}
& 2019 & VIS & CNN-based \\

Lightweight YOLOv5~\citep{yang2022lightweight}
& 2022 & VIS & YOLOv5 \\

DP-SSD~\citep{wang2022ship}
& 2022 & VIS & SSD \\

CG-YOLOv8~\citep{gong2025real}
& 2025 & VIS & YOLOv8 \\

IRS-DETR~\citep{zheng2026irs}
& 2026 & IR & RT-DETR \\

\bottomrule[1.1pt]
\end{tabular}
\label{tab:representative_detection_methods}
\end{table}

\subsubsection{Object Detection}
\begin{figure}
	\centering
	\includegraphics[width=\linewidth]{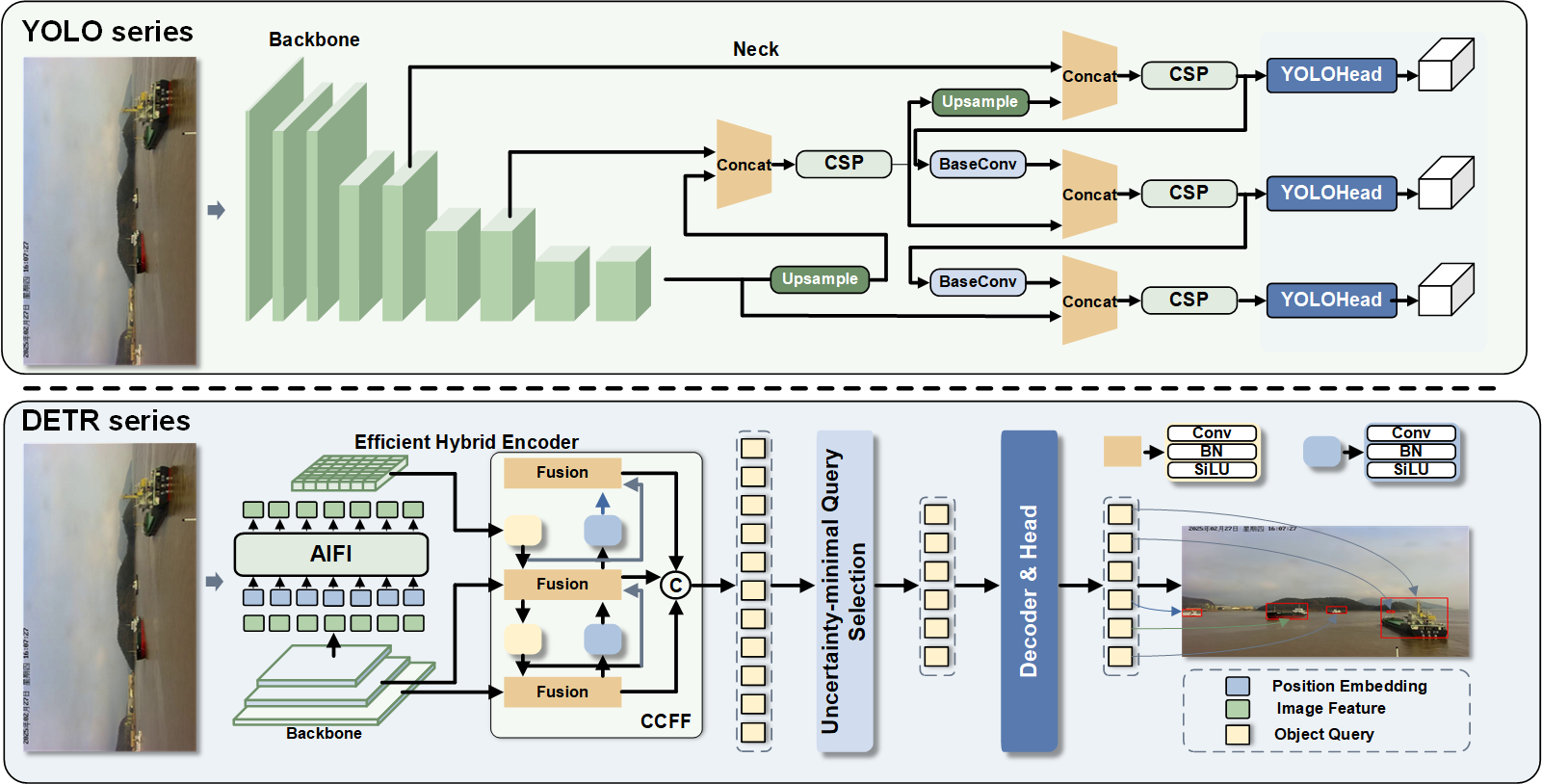}
	\caption{\textbf{Representative YOLO- and DETR-based frameworks for maritime object detection.} The upper branch illustrates multi-scale feature extraction and aggregation followed by YOLO detection heads. The lower branch illustrates a DETR-based architecture with a hybrid encoder, uncertainty-minimal query selection, and a transformer decoder for object classification and localization.}
	\label{fig:Figure_OD}
\end{figure}

Visible-light imagery provides rich texture, color, and structural cues for object recognition and localization, whereas infrared imaging captures thermal radiation independently of ambient illumination and therefore offers a complementary modality for maritime navigation. Nevertheless, both modalities remain challenging in complex maritime conditions. Existing studies can broadly be grouped into complex-scene-aware, scale-aware, and lightweight methods. Representative approaches are summarized in \cref{tab:representative_detection_methods}. Currently, most maritime object detection methods are developed based on the YOLO and DETR architectures, whose basic frameworks are illustrated in \cref{fig:Figure_OD}.

\paragraph{\textbf{Complex-scene-aware methods.}}
Maritime object detection is performed in highly variable conditions involving adverse weather, sea-surface interference, dense target distributions, and occlusions. Existing methods address these challenges through data modeling, feature representation, and detection-framework design. For visible-light imagery, Weather-OD trains specialized detectors for different weather conditions and dynamically selects the appropriate model according to the current environment~\citep{chen2024weather}. For foggy scenes, \citet{wang2024mdd} combined defogging, exposure adjustment, tone mapping, contrast enhancement, and sharpening with CNN-based parameter learning to improve image quality and detection robustness. \citet{tian2024fregnet} introduced a Feature Representation Enhancement module and a graph convolutional network (GCN)-based Combiner to fuse global semantic and local detailed key points under illumination noise and cloud interference. RetinaMFANet redesigns RetinaNet with multi-scale feature aggregation to improve representation under low illumination~\citep{ma2022improved}, while \citet{tian2025coarse} developed a coarse-to-fine dual-branch network with peer-to-peer communication to progressively localize candidate ship regions and refine contour representations.

Infrared imaging is particularly useful in low-light and nighttime conditions, but limited texture and structural cues make maritime infrared detection difficult in dynamic sea environments. \citet{zhang2024tspnet} exploited temporal differences between targets and wave clutter using accumulated energy and motion-difference maps. \citet{feng2024meta} combined meta-learning and domain adaptation to learn semantically meaningful domain-invariant features, while \citet{zhang2024infrared} incorporated dilated convolution into an attention module to enlarge the receptive field and strengthen semantic representation. To reduce missed detections and false alarms caused by densely distributed targets and occlusion, \citet{guo2024yolo} incorporated a Kolmogorov-Arnold Network (KAN)~\citep{liu2024kan} and shifted-window multi-head self-attention into YOLOv10~\citep{wang2024yolov10}, forming YOLO-IRS. \citet{ma2026spatio} further proposed a Spatio-Semantic Enhanced Cascade Swin Detection Network that combines structural-tensor-based spatiotemporal enhancement with dynamic context-aware semantic attention.

%
\paragraph{\textbf{Scale-aware methods.}}
Differences in observation distance, sensor altitude, and viewing perspective produce substantial scale variations in maritime imagery, while distant small targets can be easily obscured by sea clutter. Feature pyramid architectures~\citep{lin2017feature,ghiasi2019fpn,tan2020efficientdet} are therefore widely used to integrate multi-level features. \citet{zhang2023lightweight} combined multi-scale feature extraction, coordinate convolution, and a dynamic head to support multi-scale detection while maintaining a lightweight design. \citet{yu2024yolo} introduced multi-scale cross-axis attention to capture long-range contextual dependencies, whereas \citet{shao2022multi} used deformable convolutions together with DIoU loss and SNMS to accommodate geometric variations and improve candidate-box selection in dense scenes. 

With the success of DETR-based frameworks~\citep{zhu2020deformable,zhang2022dino,zhao2024detrs}, Transformer architectures have increasingly been adapted to maritime multi-scale detection. RTDETR-MARD combines hierarchical multi-scale fusion and feature aggregation for floating-debris detection by USVs~\citep{sun2025rtdetr}. USVRT-DETR integrates multi-kernel cascading, channel progressive compression, and a small-object-oriented feature pyramid with a P2 layer, SPDConv, and COK-Fusion~\citep{zhang2025usvrt}. For tiny boats, \citet{guo2023thfe} proposed Triple-Hierarchy Feature Enhancement (THFE) to reinforce semantic representations across network layers, while \citet{chen2020deep} combined a Gaussian Mixture Wasserstein GAN with Gradient Penalty and a CNN detector to generate additional small-ship samples and enrich training data. 

Multi-scale detection is particularly challenging in infrared imagery because wave clutter, thermal reflections, and dynamic backgrounds further reduce target contrast and boundary clarity. GLAF-DETR introduces Global-Local Adaptive Fusion and dynamic attention-guided multi-scale fusion into RT-DETR to enhance sparse and small targets~\citep{zhang2025glaf}. \citet{wu2023sarfb} used cascaded asymmetric convolutions to enlarge the receptive field with fewer parameters. \citet{sun2025irsd} combined a Hierarchical Multi-Kernel Convolution Network with a Dynamic Cross-Scale Feature Pyramid Network to integrate low-level details and high-level semantics. \citet{wang2026fgom} further proposed an infrared far-shore object-focusing framework incorporating Feature Re-parameterization, the Coordinate Attention Golden Feature Pyramid Network~\citep{wang2023gold}, and a Multi-Scale Stacked Network to improve multi-scale feature extraction and long-range modeling.
\paragraph{\textbf{Lightweight methods.}}
MSV platforms are constrained by shipboard computation, power consumption, and real-time requirements, making lightweight detection important for practical deployment. Common strategies include efficient architectures and operators such as depthwise separable convolutions~\citep{howard2017mobilenets,han2020ghostnet,ding2024unireplknet} and lightweight backbones~\citep{howard2019searching,tan2019efficientnet,chen2023run}, as well as model compression through pruning~\citep{ye2023performance,cheng2017survey} and knowledge distillation~\citep{chen2017learning,peng2025d,chang2023detrdistill}. 

For maritime detection, \citet{gong2025real} introduced depthwise separable convolutions, a P6 detection layer, and CARAFE-based lightweight upsampling into YOLOv8 to balance efficiency and multi-scale perception. \citet{yang2022lightweight} combined an improved ShuffleNetV2 backbone with attention mechanisms and depthwise separable convolutions and redesigned the neck for efficient feature extraction. \citet{wang2022ship} developed a lightweight detector evaluated in terms of computational time, frame rate, and recognition accuracy. \citet{shao2019saliency} combined deep features, saliency maps, and coastline priors to improve ship detection in complex maritime scenes. For infrared imagery, \citet{zheng2026irs} proposed a lightweight Transformer detector with a dual-branch backbone, Pinwheel-Shaped Convolution for downsampling and small-target preservation, and Matchability-Aware Loss (MAL) for dense occlusions.

\subsubsection{Object Tracking}

Object tracking continuously localizes targets across consecutive frames by modeling their appearance and motion, thereby estimating temporal trajectories. Reliable maritime tracking remains challenging because of appearance variations, dynamic backgrounds, platform motion, and frequent occlusions. Event-based sensing has also been explored for object tracking by exploiting its high temporal resolution and complementary spatio-temporal information~\citep{deng2025had}. Depending on whether an independent detector is required during inference, existing methods can generally be divided into tracking-by-detection and detection-free tracking. Tracking-by-detection first detects ships in individual frames and then associates instances using motion prediction, appearance similarity, or spatial matching. Detection-free methods generally initialize a target in the first frame and directly estimate its subsequent location through template matching, feature interaction, or temporal modeling.

\begin{table}[!t]
\caption{\textbf{Representative maritime object tracking methods.}}
\centering
\footnotesize
\setlength{\tabcolsep}{4pt}
\begin{tabular}{lc|ll}
\toprule[1.1pt]
Method & Year & Tracking paradigm & Base framework \\
\midrule

\multicolumn{4}{l}{\textit{Tracking-by-detection methods}} \\

DeepSORT~\citep{wojke2017simple}
& 2017
& Motion-appearance association
& SORT and deep Re-ID \\

YOLOv4-UKF~\citep{hong2022research}
& 2022
& Detection and motion
& YOLOv4 and UKF \\

ORB-and-size tracker~\citep{park2022object}
& 2022
& Spatial-feature association
& IoU and ORB \\

MaritimeMOT~\citep{shao2024robust}
& 2024
& Maritime target association
& YOLOv7-based MOT \\

Sea-IoUTracker~\citep{guo2024sea}
& 2024
& Buffered-IoU association
& SORT \\

VTM-YS~\citep{hao2025lightweight}
& 2025
& Motion-appearance association
& StrongSORT and OSNet \\

TIM-DETR MOT~\citep{ma2026multi}
& 2026
& Detection, association, and Re-ID
& TIM-DETR and Siamese network \\

\midrule
\multicolumn{4}{l}{\textit{Detection-free tracking methods}} \\

SiamFPN~\citep{shan2020siamfpn}
& 2021
& Siamese tracking
& Siamese network and FPN \\

MGPN~\citep{yang2023coastal}
& 2023
& Memory-guided tracking
& Siamese network and memory mechanism \\

Online Siamese-RPN~\citep{hu2023robust}
& 2023
& Online Siamese tracking
& Siamese-RPN and UpdateNet \\

\bottomrule[1.1pt]
\end{tabular}
\label{tab:representative_tracking_methods}
\end{table}

To assess tracker applicability in maritime environments, \citet{din2025benchmarking} developed a closed-loop framework integrating visual tracking with USV motion control and evaluated Siamese- and Transformer-based trackers in simulated and real scenarios, demonstrating the robustness of SeqTrack under challenging conditions such as low visibility. \citet{bakht2025mvtd} introduced MVTD, a large-scale maritime tracking benchmark containing 182 high-resolution videos and approximately 150,000 annotated frames, and evaluated 14 state-of-the-art trackers, revealing performance degradation of general-purpose trackers in maritime scenes and the benefits of domain-specific fine-tuning. Representative methods are summarized in \cref{tab:representative_tracking_methods}.

\paragraph{\textbf{Tracking-by-detection methods.}}

DeepSORT~\citep{wojke2017simple} is a representative tracking-by-detection framework that extends SORT (\textit{i.e.}, simple online and realtime tracking) with deep appearance descriptors. It combines Kalman-filter-based motion prediction with deep appearance similarity to improve association across frames and has also been widely applied to ship tracking. \citet{hong2022research} combined YOLOv4~\citep{bochkovskiy2020yolov4} with an Unscented Kalman Filter (UKF) for real-time multi-ship tracking. \citet{park2022object} combined IoU matching with ORB-and-size matching to handle rapid changes in target position and appearance. 

For more complex maritime scenarios, \citet{shao2024robust} combined YOLOv7 with shipboard-camera motion compensation, a maritime dynamic-spatial-information-based IoU metric, and progressive refinement cascade matching. Sea-IoUTracker~\citep{guo2024sea} used SORT with Buffered-IoU at two buffer scales and motion prediction to recover associations when direct adjacent-frame matching fails. To reduce computational complexity and identity switches under occlusion, \citet{hao2025lightweight} integrated OSNet~\citep{zhou2019omni} into StrongSORT for Re-ID feature extraction and used an NSA Kalman filter for trajectory prediction. For prolonged occlusions, \citet{ma2026multi} combined TIM-DETR detection and temporal feature preservation with Siamese appearance similarity, IoU-based motion constraints, and historical-feature backtracking to re-identify vessels and reduce trajectory fragmentation.
\paragraph{\textbf{Detection-free methods.}}
Detection-free tracking does not invoke an independent detector in every frame. Siamese tracking is a representative paradigm that reformulates tracking as similarity matching: shared-weight branches extract features from template and search images, and cross-correlation produces target responses. This design provides efficient sample utilization and a favorable accuracy-speed trade-off, although it can remain sensitive to severe inter-object occlusion. 

To address occlusion, scale variation, and motion blur, \citet{yang2023coastal} proposed the Memory-Guided Perception Network (MGPN), in which a Dynamic Memory Mechanism stores historical target features and a Hierarchical Context-Aware Module integrates global and local multi-scale information. \citet{hu2023robust} combined a Siamese network with a region proposal network, fused offline Siamese and online classifier scores, and introduced an occlusion-aware strategy that suspends template updates and activates global search when occlusion occurs; UpdateNet further alleviates template degradation. \citet{shan2020siamfpn} proposed SiamFPN, combining a ResNet-50 feature pyramid, a Siamese architecture, multiple region proposal networks, and historical-information-guided proposal selection for accurate and real-time maritime tracking.

\subsection{Navigable Area Segmentation}
Unlike object detection, which localizes and classifies discrete objects, shipboard semantic segmentation assigns pixel-level labels to the visual scene. It can identify navigable regions and obstacles and thereby provide environmental information for visual navigation and path planning. According to the target categories, maritime segmentation can broadly be divided into navigable-water segmentation and obstacle segmentation. The former focuses on the navigable region, whereas the latter jointly segments obstacles (\textit{e.g.}, ships, buoys, piers, and land), sky, and navigable water.

\subsubsection{Navigable Waters Segmentation}
Navigable-waters segmentation directly identifies navigable regions from perceptual data, providing scene information for path planning and external navigation constraints~\citep{kim2023navigable}. A major challenge is the visual similarity between water reflections and surrounding objects. To address this issue, \citet{zhou2022collision} constructed an ideal affinity matrix from manually annotated water and background regions and trained a contextual-prior module to model intra-region similarity and inter-region differences, improving discrimination between shoreline scenes and their reflections. \citet{ding2024research} introduced explicit water-boundary prediction as an auxiliary task and constraint on navigable-area segmentation to reduce confusion between water and land. \citet{zhou2024efficient} further exploited the correlation between image-row position and the spatial distribution of sky and navigable water, incorporating sine-encoded row-position embeddings together with water-boundary constraints to reduce sky-water confusion. 

Water reflections strongly affect cameras but have a substantially different impact on LiDAR. Accordingly, several studies have fused LiDAR point clouds with visual information~\citep{gao2022camera,grini2024fusedwss} or used LiDAR alone~\citep{shan2021lidar} for navigable-area segmentation. Because most laser energy incident on the water surface is absorbed or specularly reflected, the resulting point clouds are dominated by returns from land and other solid structures. Cross-modal methods can therefore use the three-dimensional geometric structure provided by LiDAR as auxiliary information to reduce reflection-induced visual ambiguity, while $360^\circ$ LiDAR-based approaches can also infer navigable regions from this geometric characteristic~\citep{shan2021lidar}. 

\subsubsection{Obstacle Segmentation}
Perceiving only the navigable region is insufficient for higher-level functions such as navigation decision-making, which require a more complete representation of the surrounding environment. Obstacle segmentation therefore jointly distinguishes obstacles, sky, and navigable water. Using the MaSTr1325~\citep{bovcon2019mastr1325} and MODS~\citep{bovcon2021mods} datasets, Bovcon \textit{et al.} evaluated general semantic segmentation models including U-Net~\citep{ronneberger2015u}, PSPNet~\citep{zhao2017pyramid}, and DeepLab v2~\citep{chen2017deeplab}, showing that illumination variations and water reflections can cause segmentation errors and inaccurate object contours.

To improve robustness to vessel motion, \citet{bovcon2018stereo} incorporated inertial measurement unit (IMU) data into probabilistic semantic segmentation. Roll and pitch measurements are used to estimate the horizon and adapt the conditional prior distribution and Gaussian hyperprior, improving segmentation under severe vessel sway. Because handcrafted graphical models remain limited under reflections, highlights, spray, and wakes, \citet{bovcon2020water,bovcon2021wasr} developed WaSR, which combines deep visual features with shipboard IMU information through an asymmetric encoder-decoder architecture and introduces semantic separation loss to improve feature-space separation among semantic components. \citet{vzust2022temporal} further proposed WaSR-T by incorporating visual observations from the current and previous $T$ frames and extracting temporal context to reduce ambiguity from temporally unstable reflections and highlights. 

Beyond reflection interference, small-object segmentation and false positives near the water-sky boundary remain challenging. \citet{chen2021wodis} proposed WODIS, an encoder-decoder network that extracts multi-rate high-level features, uses an attention-refinement module based on global average and max pooling for sea-sky-line regions, and integrates multidimensional features through a feature-fusion module. \citet{feng2024lightweight} developed a dual-branch architecture consisting of a dilated-convolution semantic branch~\citep{yu2015multi} and a detail branch centered on an embedded feature-refinement module. Directional feature refinement and asymmetric convolution emphasize stable structural components while suppressing high-frequency dynamic interference such as water reflections, and the two branches are progressively fused during inference. 

Boundary accuracy constitutes another major challenge. \citet{zhang2024boundary} introduced separate boundary-extraction and boundary-enhancement modules to explicitly fuse boundary cues with global semantic features, improving pixel classification around object contours while preserving regional segmentation. \citet{zhang2025scene} proposed a multi-level boundary-semantic reinforcement mechanism that combines explicit boundary supervision, Coordinate Attention for spatial localization, and a global-local boundary-aware discriminator based on shape context to adaptively emphasize geometrically inconsistent regions. Camera-based surround-view bird's-eye view (BEV) segmentation has attracted attention in autonomous driving, but its development for water-surface environments remains limited by the scarcity of corresponding BEV datasets. Maritime BEV segmentation also faces vessel sway and reflection interference. To support this direction, \citet{xu2024surround} introduced the WSBEV dataset using a hybrid configuration of pinhole and fisheye cameras with different focal lengths. The Hybrid-BEV was then proposed, which was essentially a water-surface BEV segmentation network supporting heterogeneous camera inputs. 

Overall, recent work has improved the accuracy and stability of maritime scene segmentation. However, environmental interference, small-object segmentation, and boundary accuracy remain important challenges. The attention mechanisms, global-local feature modeling, temporal modeling, and multi-modal data fusion provide potential directions for improving robustness, while lightweight and high-performance models remain important for practical MSV deployment.

%
%
\section{Vision-Centered Multi-Sensor Fusion}\label{sec:datafusion}
%
To further improve maritime situational awareness, MSVs can exploit complementary information from multiple shipboard sensors. Vision-centered multi-sensor fusion integrates visual observations with other sensing modalities to improve environmental perception in non-ideal navigation conditions. This section reviews representative fusion strategies, including infrared-visible image fusion and the fusion of vision with radar, LiDAR, AIS, and other shipboard sensors, with particular emphasis on navigational safety in restricted visibility.%
\subsection{Infrared and Visible Image Fusion}
Both infrared and visible-light cameras have been widely used for surface-vessel perception~\citep{bloisi2016enhancing}. Visible imagery provides rich appearance information but is vulnerable to adverse illumination and weather, whereas infrared imagery captures thermal radiation but generally has lower spatial resolution and a more limited field of view (FOV). The substantial modality gap between visible and infrared imagery has also motivated cross-modal representation learning to improve their correspondence~\citep{GrayscaleEnhancement}. Infrared-visible fusion therefore seeks to preserve complementary thermal and appearance information, improving the robustness and accuracy of vessel recognition~\citep{el2023novel}. For example, \citet{lu2023fusion2fusion} proposed an attention-driven framework consisting of feature extraction, fusion, and reconstruction modules to improve navigational-environment perception. Calibrated and aligned infrared and visible images have also been used to enhance shipboard visual perception~\citep{zhang2022infrared}. \citet{farahnakian2019visible} incorporated fused imagery into a CNN-based RetinaNet detector to improve MSV detection in different weather and illumination conditions, and further investigated pixel-, feature-, and decision-level fusion strategies for maritime object detection~\citep{farahnakian2020deep}. The VAIS benchmark, containing more than $1000$ paired infrared and visible images across six categories, was constructed to evaluate vessel recognition~\citep{zhang2015vais}. More comprehensive discussions of infrared-visible image fusion can be found in~\citep{ma2019infrared,zhang2023visible}.
\subsection{Radar and Vision Fusion}
Visual sensors provide rich semantic and categorical information but have limited capability to directly measure relative distance, velocity, and course between a target and the own ship. Navigation radar on large MSVs and 4D millimeter-wave (mmWave) radar on smaller MSVs can provide complementary geometric and kinematic information, while maintaining all-day and all-weather sensing capabilities under low illumination and adverse weather~\citep{yao2023radar,yao2025exploring}. Radar-vision fusion can therefore jointly exploit visual semantics and target motion information for more comprehensive maritime situational awareness. Existing studies mainly consider navigation-radar-vision fusion and 4D-mmWave-radar-vision fusion.

\subsubsection{Navigation Radar and Vision Fusion}
Navigation radar and cameras differ substantially in sensing geometry and data representation, creating a pronounced cross-modal gap. Navigation radar measures range, velocity, and course in polar coordinates and commonly represents observations in BEV, whereas cameras map the three-dimensional environment onto a two-dimensional perspective image~\citep{shi2025radar}. Cross-modal matching and alignment are therefore central to navigation-radar-vision fusion. 

Early methods primarily relied on static geometric alignment. \citet{liu2021hybrid} and \citet{clunie2021development} used calibrated extrinsic parameters and coordinate transformations to project radar detections onto the image plane. Although straightforward, such approaches are sensitive to calibration drift caused by vessel pitching and long-term vibration. Adaptive and probabilistic matching strategies have subsequently been developed to reduce this dependence. \citet{farahnakian2018object} introduced cross-modal consistency checks to filter redundant visual proposals using radar-derived spatial constraints. \citet{cormack2019joint} proposed a joint registration and fusion framework using probabilistic data association to accommodate measurement uncertainty and dynamic sensor offsets. \citet{kim2021robust} improved matching robustness by optimizing geometric parameters, reducing dependence on precise initial calibration. AnytimeFusion~\citep{shin2024anytimefusion} further employed particle swarm optimization for fully parameterized automatic alignment in dynamic maritime environments. 

More recent studies have shifted toward learned cross-modal association. \citet{xu2025multi} introduced a Transformer-based fusion module that uses cross-attention to associate radar observations with visual semantics in a latent feature space, reducing representational discrepancies between the two modalities. For small water-surface targets, where radar echoes are sparse and strongly affected by multipath clutter, \citet{wei2024small} proposed multi-frame radar feature accumulation to improve temporal density and continuity, followed by image-guided radar feature enhancement that uses high-resolution visual semantics to refine sparse radar representations.
\subsubsection{4D mmWave Radar and Vision Fusion}
\begin{figure}
    \centering
    \includegraphics[width=\linewidth]{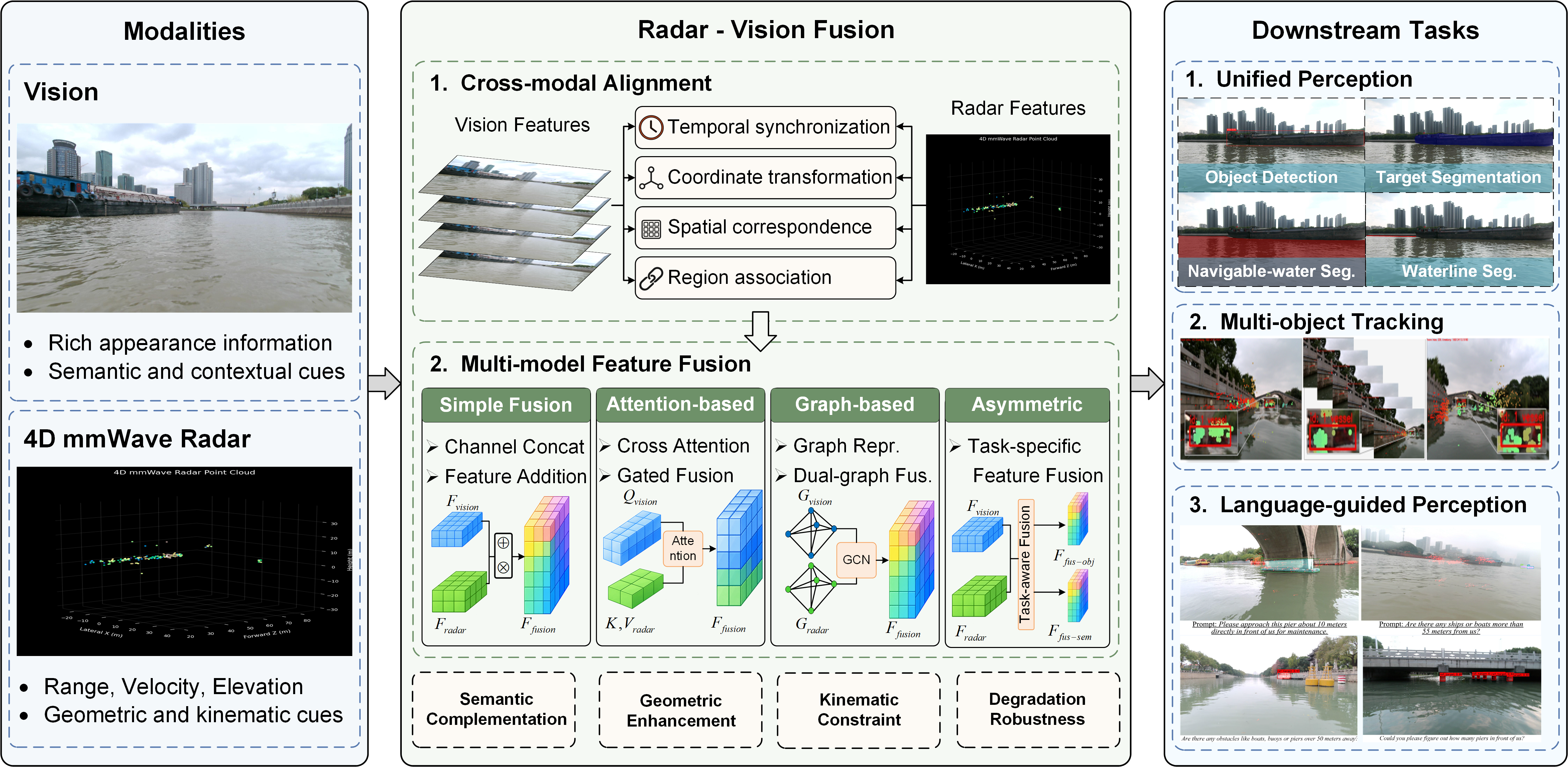}
    \caption{\textbf{Schematic overview of 4D mmWave radar-vision fusion.} Vision and 4D mmWave radar provide complementary semantic, geometric, and kinematic information. Cross-modal alignment establishes temporal, spatial, and regional correspondences between the two modalities, followed by radar-vision feature fusion using simple, attention-based, graph-based, or asymmetric fusion strategies. The fused features support downstream maritime perception tasks, including unified perception, multi-object tracking, and language-guided perception.}
    \label{fig:4D_mmWave_radar_vision_fusion}
\end{figure}
Compared with navigation radar, 4D mmWave radar additionally measures elevation, extending perception from a two-dimensional plane to three-dimensional space, as illustrated in \cref{fig:4D_mmWave_radar_vision_fusion}. Benefiting from this enhanced spatial sensing capability, 4D-mmWave-radar-vision fusion has been increasingly explored in autonomous driving for tasks such as 3D object detection and tracking, simultaneous localization and mapping, and BEV perception under diverse weather and illumination conditions~\citep{fan20244d}. In contrast, the application of 4D-mmWave-radar-vision perception to maritime environments has emerged more recently and remains less extensively studied. Even so, recent advances in dedicated datasets, perception frameworks, and task-specific methods have begun to establish a solid foundation for further research on 4D-mmWave-radar-vision perception in maritime scenarios.

Representative datasets, such as WaterScenes~\citep{yao2024waterscenes} and USVInland~\citep{cheng2021we}, have been constructed for 4D-mmWave-radar-vision perception under diverse illumination and weather conditions. These datasets have supported studies on challenging tasks, \textit{e.g.}, small maritime object detection~\citep{cheng2021robust,he2024millimeter}. 

Several studies have subsequently explored unified maritime perception. Achelous and Achelous++~\citep{guan2023achelous,guan2023achelous++} introduced multi-task panoptic perception frameworks that jointly address heterogeneous tasks including object segmentation, navigable-area segmentation, object detection, waterline segmentation, and radar point-cloud segmentation. To improve cross-modal alignment and robustness against water-surface interference, Mask-VRDet~\citep{guan2024mask} introduced a dual-graph fusion framework to model non-Euclidean relationships between visual and radar features, together with a clutter-point-removal strategy. To accommodate different modality requirements across tasks, \citet{guan2024asy} further proposed an asymmetric fair fusion module that represents image and radar observations as irregular superpixel sets and interacts with them in a shared cross-modal feature space. Aiming at the issues of asynchronous modality degradation and the resulting decline in perception performance that can arise when using 4D mmWave radar and vision sensors in dynamic maritime environments, \citet{song2026uncertainty} proposed an uncertainty-aware radar-vision fusion method. By explicitly modeling the uncertainty inherent in visual and radar features, this method enables modality reliability allocation and the isolation of degradation noise at the feature level, thereby enhancing the robustness of fused perception in complex, dynamic environments.

Beyond detection and segmentation, radar-vision fusion has also been extended to tracking. \citet{yao2025usvtrack} constructed USVTrack, a comprehensive 4D-mmWave-radar-vision tracking dataset for inland waterways, and proposed Radar-Camera Matching (RCM). RCM uses radar-derived distance and velocity as kinematic constraints for cross-frame target association, reducing identity ambiguity during visual occlusion. 

Recent studies have further extended radar-vision perception toward language-guided interaction. \citet{guan2025watervg} introduced WaterVG, a waterway visual-grounding dataset containing more than 11,000 samples and 35,000 referred targets, with textual descriptions covering semantic information and numerical kinematic attributes such as distance and speed. The associated Potamoi model performs referring-expression comprehension and segmentation, enabling language-guided target localization and segmentation. NanoMVG~\citep{guan2025nanomvg} subsequently introduced a low-power multi-task framework optimized for edge deployment while maintaining localization and segmentation capability. For autonomous navigation, occupancy modeling provides an important bridge between perception and path planning. \citet{xu2024real} proposed a real-time 4D-mmWave-radar-camera fusion network for BEV occupancy prediction. A BEV spatial Transformer projects perspective-view image features into a radar-aligned BEV space, enabling two-dimensional occupancy mapping while suppressing water-surface clutter and noise. 

These studies demonstrate the complementary roles of vision and 4D mmWave radar in maritime perception. Vision provides rich appearance and semantic information, whereas 4D mmWave radar contributes direct range, radial-velocity, and elevation measurements that provide additional geometric and kinematic constraints, particularly when visual observations are degraded. Their fusion therefore improves spatial localization, target association, tracking, and scene understanding under dynamic maritime conditions. However, current studies have predominantly focused on short-range to medium-range MSV perception, largely because the sensing range of existing 4D mmWave radar constrains long-range maritime operation. The existing radar-vision fusion methods extend to longer-range perception once radar sensing delivers reliable measurements at the required distances.
\begin{figure}
	\centering
	\includegraphics[width=\linewidth]{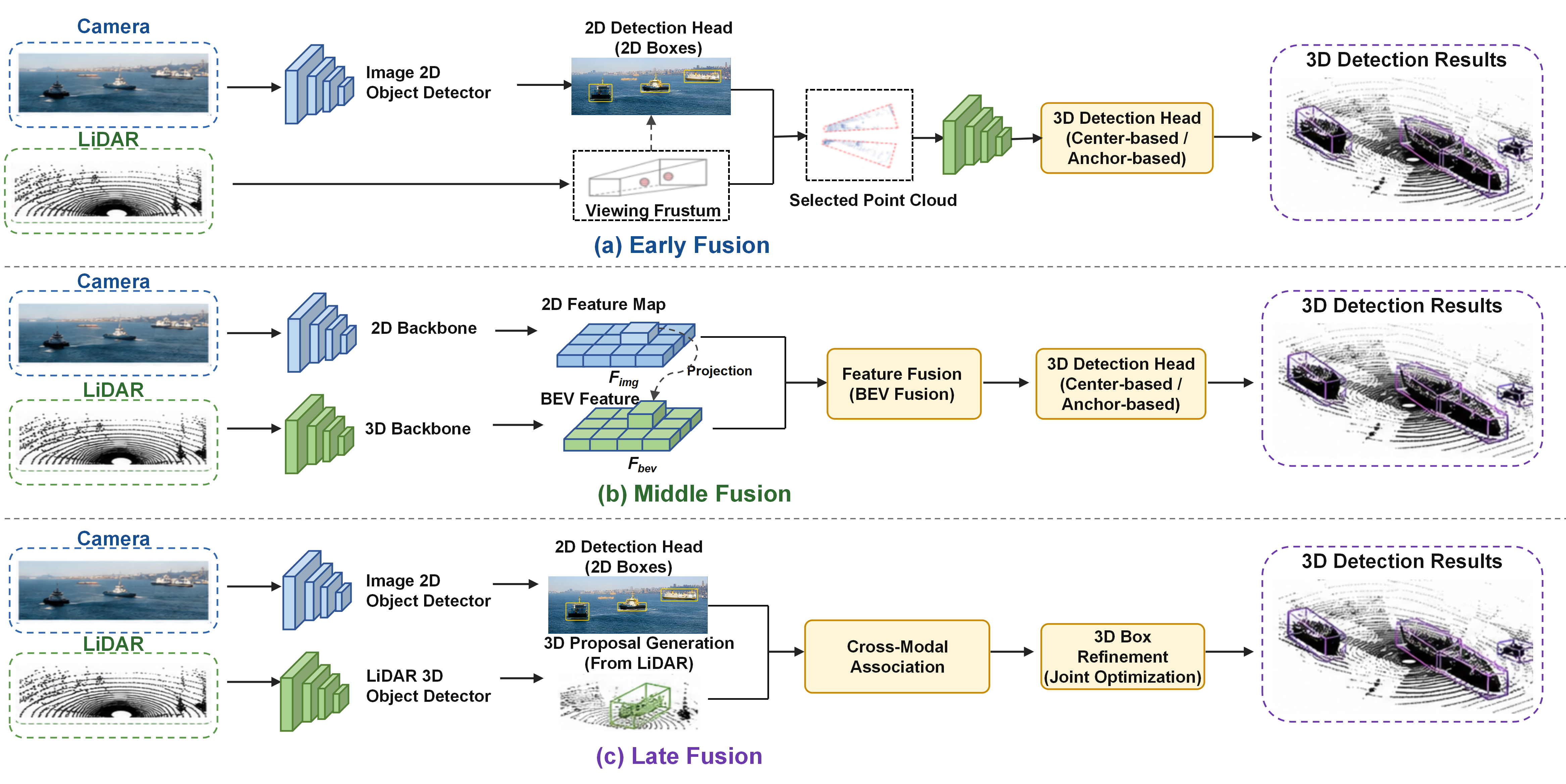}
	\caption{\textbf{Representative LiDAR--camera fusion paradigms for maritime 3D object detection.} (a) Image detections define viewing frustums for selecting LiDAR points before 3D detection. (b) Image and LiDAR features are aligned and fused in a bird's-eye-view representation. (c) Independently generated image detections and LiDAR proposals are associated across modalities and jointly refined to produce 3D bounding boxes.}
	\label{fig:LVFusion}
\end{figure}
\subsection{LiDAR and Vision Fusion}

%
%
LiDAR provides accurate geometric and depth information, whereas cameras capture rich appearance and semantic information, making the two modalities complementary for three-dimensional environmental perception. In autonomous driving, LiDAR-vision fusion is commonly categorized as Early, Middle, and Late fusion according to the stage at which cross-modal interaction occurs, shown in \cref{fig:LVFusion}. These paradigms also provide useful methodological references for maritime perception.
\begin{itemize}
	\item Early fusion combines multimodal observations during preprocessing or at the input stage, typically through spatial alignment. This strategy preserves raw cross-modal information and facilitates direct modeling of geometric-appearance relationships. PointPainting~\citep{vora2020pointpainting} and PointAugmenting~\citep{wang2021pointaugmenting}, for example, project point clouds onto the image plane or construct unified three-dimensional representations to enrich point-cloud features with visual information.
	\item Late fusion combines independently generated detection or recognition outputs. Frustum-PointNets~\citep{qi2018frustum} first obtain image-based 2D detections and project the corresponding regions of interest (RoIs) into 3D frustums, within which PointNet-like networks estimate 3D bounding boxes. RoarNet~\citep{shin2019roarnet} first estimates geometrically feasible 3D candidate regions from monocular images and subsequently refines them using a 3D detector. Frustum-PointPillars~\citep{paigwar2021frustum} combines visual detections with PointPillars~\citep{lang2019pointpillars} representations to support localization under sparse point clouds.
	\item Middle fusion has become a major paradigm because it enables intermediate semantic features from different modalities to interact directly. Early approaches often projected point clouds into multiple 2D views. MV3D~\citep{chen2017multi}, for example, projects point clouds into BEV and front-view representations and fuses image, BEV, and front-view features through RoI pooling.
\end{itemize}

More recent methods seek to preserve three-dimensional spatial information and learn adaptive cross-modal alignment. MMF~\citep{liang2019multi} uses continuous fusion layers to map image features into BEV representations; EPNet~\citep{huang2020epnet} establishes point-wise correspondences between point clouds and image features; and 3D-CVF~\citep{yoo20203d} performs cross-view mapping between voxel and image features. VPFNet~\citep{zhu2022vpfnet} fuses point-cloud voxels and stereo-image features through virtual grid points, while SFD~\citep{wu2022sparse} combines voxel and pseudo-point-cloud representations. AutoAlign~\citep{chen2022autoalign} introduces learnable alignment and cross-attention for adaptive pixel-to-voxel aggregation. BEVFusion~\citep{liu2023bevfusion} unifies multiple modalities in a shared BEV representation, whereas TransFusion~\citep{bai2022transfusion} uses Transformer decoders to initialize object queries from LiDAR features and fuse them with image features through attention. DeepFusion~\citep{li2022deepfusion} similarly converts modality-specific features into a common BEV representation for subsequent fusion. 

Compared with autonomous driving, LiDAR-camera fusion for MSVs remains relatively underexplored. Existing studies have first investigated LiDAR-only maritime perception. \citet{eustache2025marine} released a USV point-cloud dataset augmented through semi-physical simulation. For busy maritime environments, \citet{xie2024reliable} proposed a modular LiDAR-based detection framework together with Kalman-filter-based multi-object tracking to compensate for missed detections caused by motion and occlusion. \citet{guo2026lidar} combined average and attention pooling to aggregate local geometry and point intensity within pillars. \citet{ponzini2025lidar} integrated lightweight supervised classification into an unsupervised maritime detection pipeline, while \citet{xie2026uncertainty} combined attention-based pillar feature extraction with a hybrid MobileViT-CNN backbone for global representation learning. 

LiDAR-only perception nevertheless remains sensitive to waves, reflections, and other environmental noise and provides limited appearance information for object recognition and classification. LiDAR-vision fusion therefore offers a natural means of combining precise geometry with visual semantics for MSV navigation. \citet{zhang2021research} developed an integrated WAM-V-USV perception and obstacle-avoidance framework combining LiDAR and vision with water-reflection removal, image denoising, sea-sky detection, and obstacle avoidance. \citet{chen2021novel} proposed a Three-Dimensional Perception Module (PMTD) that combines camera-LiDAR perception with GPS and IMU localization to support avoidance of static and dynamic obstacles. 

For water-surface target detection and localization, \citet{wang2023water} first detects candidate targets using CornerNet-Lite, projects LiDAR points onto the image plane, and combines camera and LiDAR confidence scores to determine targets and estimate their 3D positions. An image-dominant fusion framework was further proposed \citep{zhang2024multisensor}. In particular, the visual observations served as the primary modality and LiDAR provided complementary information. The tracking-assisted detection and confidence-based association were then exploited to compensate for missed detections. To alleviate the shortage of maritime multimodal data, a multimodal dataset \citep{choi2025polaris} with image- and point-level annotations was introduced for obstacle detection and tracking, including detailed annotations for small objects. \citet{yao2025real} developed a two-stage point-cloud clustering strategy using adaptive thresholds and elliptical heat zones to suppress mixed target-background false detections, followed by saliency-based segmentation for target-contour extraction. For autonomous docking, \citet{xu2026markless} proposed a markerless berth-detection framework that combines scene semantics with camera-LiDAR fusion. It integrates a LiDAR-assisted view Transformer and fusion-attention module to generate semantic 3D point clouds and identify available docking spaces.
\subsection{AIS and Vision Fusion}

\begin{figure}
	\centering
	\includegraphics[width=\linewidth]{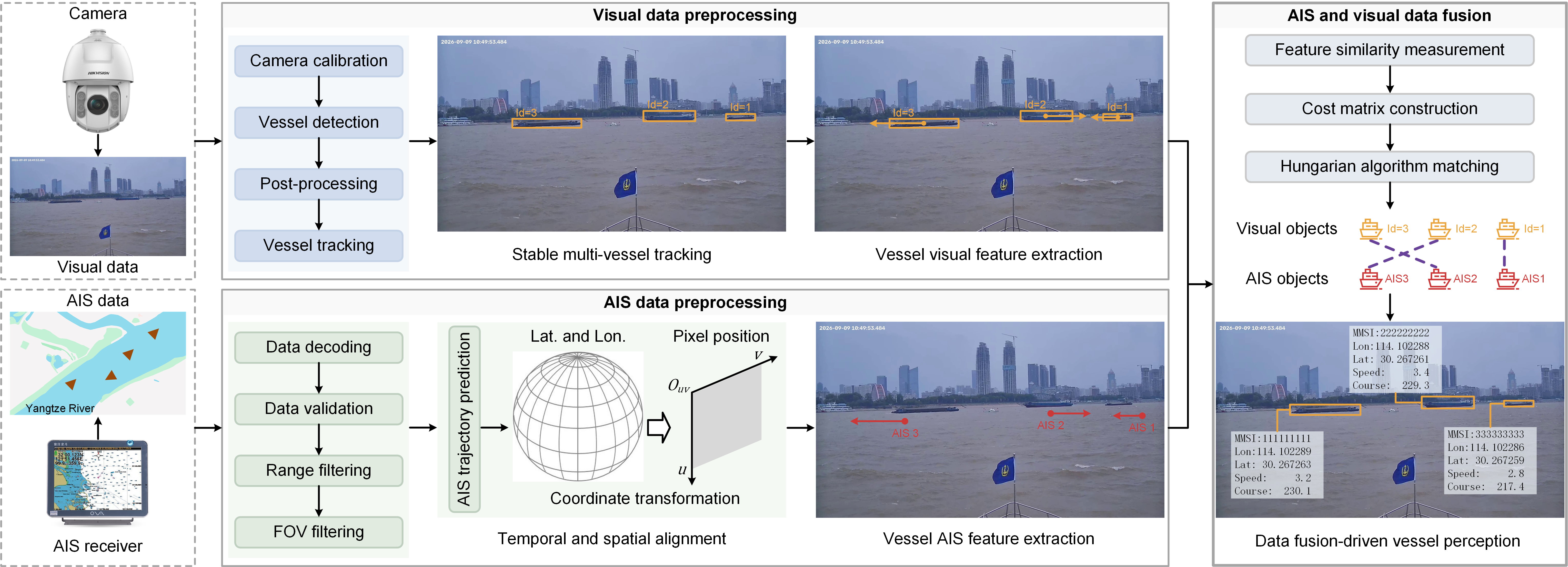}
	\caption{\textbf{Flowchart of AIS and vision data fusion.} The AIS and vision data fusion contributes to an augmented reality (AR) navigation system for intelligent surface vehicles. It improves situational awareness and decision-making capacities for safe maritime navigation under complex conditions.}
	\label{fig:AISCameraFusion}
\end{figure}

AIS provides vessel identity, position, and motion information, whereas cameras offer complementary appearance and semantic cues. Their fusion can achieve the synchronized display of the vessel's appearance and identity information, thus improving the maritime situational awareness, shown in Fig.~\ref{fig:AISCameraFusion}. \citet{liu2022intelligent} fused YOLOX-based visual detections with synchronized AIS messages using distance and angular constraints, supporting an augmented-reality (AR)-assisted intelligent navigation system. \citet{huang2023online} combined image observations with AIS messages to estimate ship dimensions in real time and reduce potential collision risk. \citet{ding2024robust} jointly employed multi-region mapping and multi-feature similarity measurement to improve AIS-vision association and vessel collision-risk assessment in inland waterways, including in low-light conditions. The pioneering AR navigation system based on AIS and vision fusion was developed by Furuno Electric Co., Ltd. More details on representative companies and their navigation systems could be found in Section~\ref{sec:navigationsystems}.

To address spatial and temporal discrepancies between AIS and camera observations, \citet{xiao2025novel} introduced regional-division projection and graph-based multi-feature fusion framework. The AIS-vision fusion has also been investigated for maritime traffic situational awareness in diverse environments~\citep{guo2023asynchronous,qu2023improving,lu2026graph}. To improve maritime traffic safety, this fusion strategy has been successfully extended to shore-based maritime surveillance~\citep{holst2026fusing,teixeira2026computer}, and UAV-based surface-vessel identification and tracking~\citep{zhao2025multi,zhao2026multi}.
\begin{figure}
	\centering
	\includegraphics[width=\linewidth]{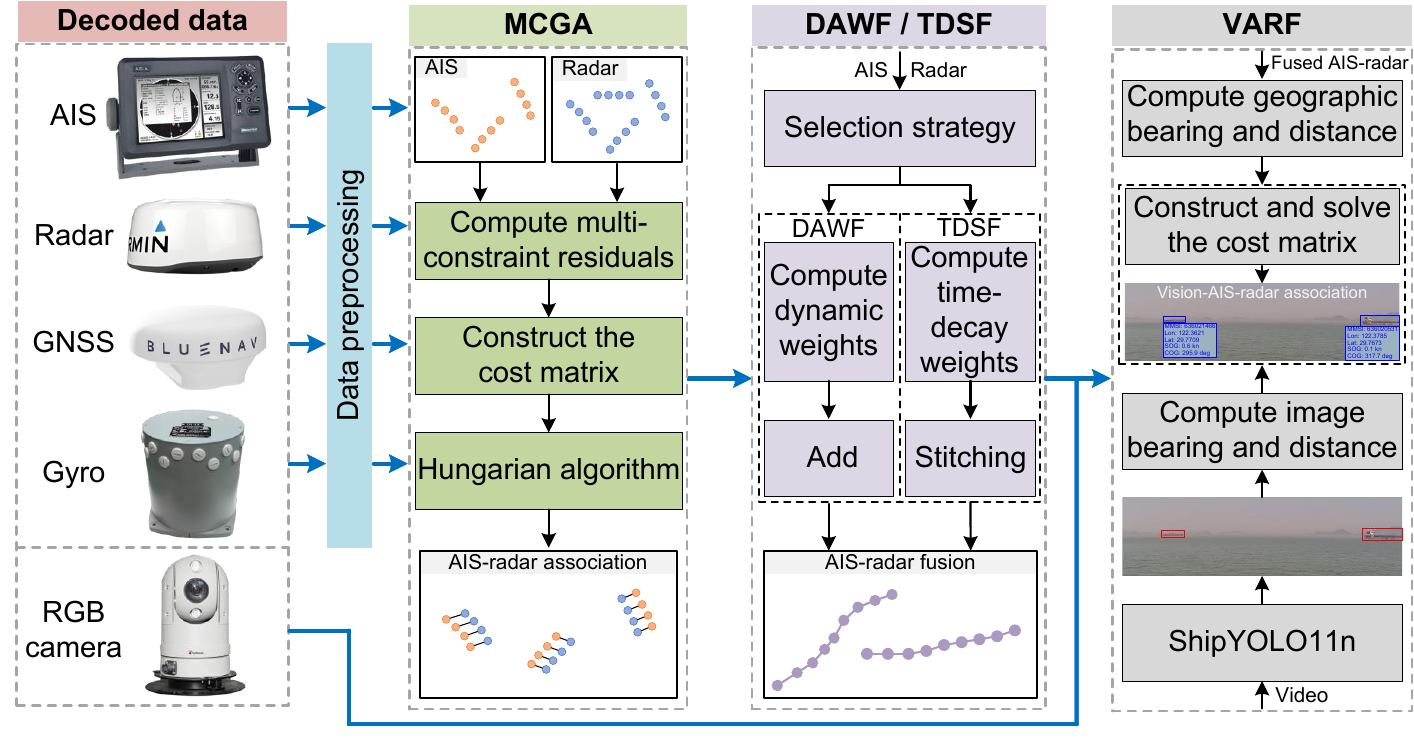}
	\caption{\textbf{Framework of vision-centered multi-sensor fusion.} In particular, gyro denotes the gyrocompass device. MCGA, DAWF/TDSF, and VARF, respectively, denote the AIS–radar data association, AIS–radar data fusion, and vision–AIS–radar data fusion methods. The selection strategy seselects the appropriate fusion method based on available observation sources. Please refer to Ref.~\citep{liu2026naveye} for more details on vision-centered multi-sensor fusion and its application in situational awareness for safe navigation.}
	\label{fig:Figure_NavEYE}
\end{figure}
%
\subsection{Other Multi-Sensor Fusion Scenarios}%
Because individual sensors exhibit complementary strengths and limitations, integrating more than two sensing modalities can further improve maritime situational awareness in dynamic and complex environments~\citep{zhang2025multi}. The Maritime RobotX Challenges~\citep{stanislas2018multimodal} have promoted the development and validation of such heterogeneous sensor-fusion systems. For example, probabilistic fusion of radar, LiDAR, and cameras has been investigated for robust target detection and classification in uncertain conditions, supporting autonomous navigation and collision avoidance. 

Several studies have developed integrated multi-sensor navigation systems. \citet{Han2020Autonomous} presented both hardware and software components of an autonomous collision-avoidance system, while subsequent field experiments demonstrated fully unmanned operation in complex coastal traffic environments~\citep{Han2023Field}. \citet{clunie2021development} combined monocular cameras with radar and LiDAR to estimate the position, dimensions, and heading of static and dynamic targets. For real-time detection and tracking, \citet{haghbayan2018efficient} proposed probabilistic data association using independent detections from radar, LiDAR, RGB cameras, and infrared cameras. Joint integrated probabilistic data association (JIPDA) and its multi-sensor extension were further developed and experimentally validated using these four sensing modalities~\citep{helgesen2022heterogeneous}. \citet{mousazadeh2018developing} fused range-finder LiDAR, stereo vision, and ultrasonic sensors through an extended Kalman filter (EKF), demonstrating robustness under obstacles and environmental disturbances including wind, waves, and currents. 

More recently, \citet{liu2026naveye} developed a NavEYE system to promote maritime situational awareness and safety navigation in complex environments, shown in~\cref{fig:Figure_NavEYE}. It is a vision-centered multi-sensor fusion system by integrating radar, RGB cameras, AIS, global navigation satellite system (GNSS), gyroscopes, and other shipboard sensors. To extend the perception range of MSVs, a new concept, \textit{i.e.}, beyond-visual-range (BVR) situational awareness~\citep{zhao2026beyond}, was proposed using the maritime Internet of Things (mIoT)-based collaborative framework. It has the capacity of detecting targets beyond the shipboard sensing coverage. 
%
%
\section{Datasets and Safety-Critical Applications}\label{sec:datasets}
Deep learning methods for intelligent vision and multi-sensor data fusion depend strongly on the scale and diversity of training data. This section first reviews representative public datasets for maritime perception and then discusses safety-critical applications of intelligent vision and vision-centered data fusion, including vessel behavior prediction, intelligent collision avoidance, and vision-centered navigation systems.

\begin{sidewaystable}[!htbp]
\caption{\small \textbf{Overview of publicly available datasets for MSVs.} Tasks: OD: object detection; LS: waterline segmentation; OT: object tracking; SS: semantic segmentation; FS: free-space segmentation; PS: panoptic segmentation; IS: instance segmentation; PP: panoptic perception.}
\centering
\footnotesize
\begin{tabular}{lc|lllrr}
\toprule[1.1pt]
Dataset & Year & Sensors & Tasks & Annotations & Classes & Frames \\
\midrule
MODD~\citep{kristan2015fast}
& 2015 & Monocular Camera & OD, LS
& 2D Box, 2D Line & 2 & 4,454 \\

MODD2~\citep{bovcon2018stereo}
& 2018 & Stereo Camera & OD, LS
& 2D Box, 2D Line & 2 & 11,675 \\

SMD~\citep{moosbauer2019benchmark}
& 2019 & Monocular Camera & OD, OT
& 2D Box & 10 & 31,653 \\

MaSTr1325~\citep{bovcon2019mastr1325}
& 2019 & Monocular Camera & SS
& 2D Pixel & 4 & 1,325 \\

MODS~\citep{bovcon2021mods}
& 2021 & Stereo Camera & OD, SS
& 2D Box, 2D Line & 3 & 24,090 \\

MID~\citep{liu2021efficient}
& 2021 & Monocular Camera & OD
& 2D Box & 2 & 2,655 \\

USVInland~\citep{cheng2021we}
& 2021 & Stereo Camera & SS, FS
& 2D Line & 1 & 700 \\

FloW~\citep{cheng2021flow}
& 2021 & Monocular Camera, 3D Radar & OD
& 2D Box & 1 & 2,000 \\

SeaSAw~\citep{kaur2022sea}
& 2022 & Monocular Camera & OD
& 2D Box & 12 & 1.9M \\

LaRS~\citep{vzust2023lars}
& 2023 & Monocular Camera & SS, PS
& 2D Line & 11 & 4,006 \\

MVDD13~\citep{wang2024marine}
& 2024 & Monocular Camera & OD
& 2D Box & 13 & 35,474 \\

WaterScenes~\citep{yao2024waterscenes}
& 2024 & Monocular Camera, 4D Radar & OD, IS, SS, PP
& 2D Box, 2D Pixel, 3D Point & 7 & 54,120 \\

OASIs~\citep{kim2024introducing}
& 2024 & Monocular Camera & SS
& 2D Pixel & 4 & - \\

MVTD~\citep{bakht2025mvtd}
& 2025 & Monocular Camera & OT
& 2D Box, Track ID & 4 & $\sim$150k \\

PoLaRIS~\citep{choi2025polaris}
& 2025 & Monocular/TIR Camera, LiDAR, Radar & OD, OT
& 2D Box, 3D Point, Track ID & 2 & $\sim$360k \\

USVTrack~\citep{yao2025usvtrack}
& 2025 & Monocular Camera, 4D Radar & OD, OT
& 2D Box, 3D Point, Track ID & 3 & 68,822 \\

MULTIAQUA~\citep{muhovivc2025multiaqua}
& 2025 & Stereo/Thermal/NIR Camera, LiDAR, Radar & SS
& 2D Pixel & 4 & 3,293 \\

SeaClips~\citep{denk2026seaclips}
& 2026 & Monocular Camera & OD
& 2D Box & 7 & 31,606 \\

WUTDet~\citep{liang2026wutdet}
& 2026 & Monocular Camera & OD
& 2D Box & 1 & 100,576 \\
\bottomrule[1.1pt]
\end{tabular}
\label{tab:datasets}
\end{sidewaystable}

\subsection{Datasets}

From the perspective of navigational safety, the development of maritime perception datasets can be broadly viewed in three stages, as summarized in \cref{tab:datasets}. The first is dominated by monocular and stereo cameras, represented by MODD~\citep{kristan2015fast}, MODD2~\citep{bovcon2018stereo}, MID~\citep{liu2021efficient}, SMD~\citep{moosbauer2019benchmark}, and MaSTr1325~\citep{bovcon2019mastr1325}. These datasets established fundamental maritime vision tasks, including horizon and waterline estimation, obstacle detection, tracking, and navigable-area segmentation. Their relatively low acquisition and annotation costs facilitated early benchmarking. However, visible imagery is readily degraded by glare, fog/haze, nighttime illumination, specular reflections, wave disturbances, and long-range small targets, while monocular and stereo vision provide limited robustness in geometric and motion perception.

The second stage is characterized by larger and more diverse RGB benchmarks, such as LaRS~\citep{vzust2023lars}, MVDD13~\citep{wang2024marine}, SeaClips~\citep{denk2026seaclips}, and WUTDet~\citep{liang2026wutdet}. Progress is reflected not only in dataset scale but also in scene diversity, category complexity, temporal structure, and evaluation protocols. LaRS~\citep{vzust2023lars} covers lakes, rivers, and seas with 19 global scene attributes and panoptic annotations containing three \textit{stuff} and eight \textit{thing} classes, extending maritime perception toward panoptic scene understanding. SeaClips~\citep{denk2026seaclips} emphasizes video-based frame-level detection, whereas WUTDet~\citep{liang2026wutdet}, with more than 100k images, provides a large-scale benchmark for dense small-ship detection in complex port and navigation environments. These datasets increasingly emphasize reliable generalization under diverse scene distributions and substantial target-scale variations.

The third stage moves toward multimodal, safety-oriented, and visibility-aware datasets. Representative examples include USVInland~\citep{cheng2021we}, FloW~\citep{cheng2021flow}, WaterScenes~\citep{yao2024waterscenes}, PoLaRIS~\citep{choi2025polaris}, USVTrack~\citep{yao2025usvtrack}, and MULTIAQUA~\citep{muhovivc2025multiaqua}. These datasets introduce complementary modalities such as millimeter-wave/4D radar, LiDAR, thermal infrared (TIR), and near-infrared (NIR), while emphasizing synchronization, extrinsic calibration, cross-modal projection, and temporal ID consistency~\citep{yao2026gif}. They reflect a transition from vision-only perception toward vision-centered data fusion, where cameras provide rich semantic and boundary information while radar, thermal infrared, and LiDAR contribute complementary geometric, structural, and sensing capabilities in restricted visibility. Accordingly, these benchmarks provide an increasingly important foundation for studying cross-modal perception and safety-relevant maritime navigation.%
\subsection{Vessel Behavior Prediction}
\begin{figure}
    \centering
    \includegraphics[width=\linewidth]{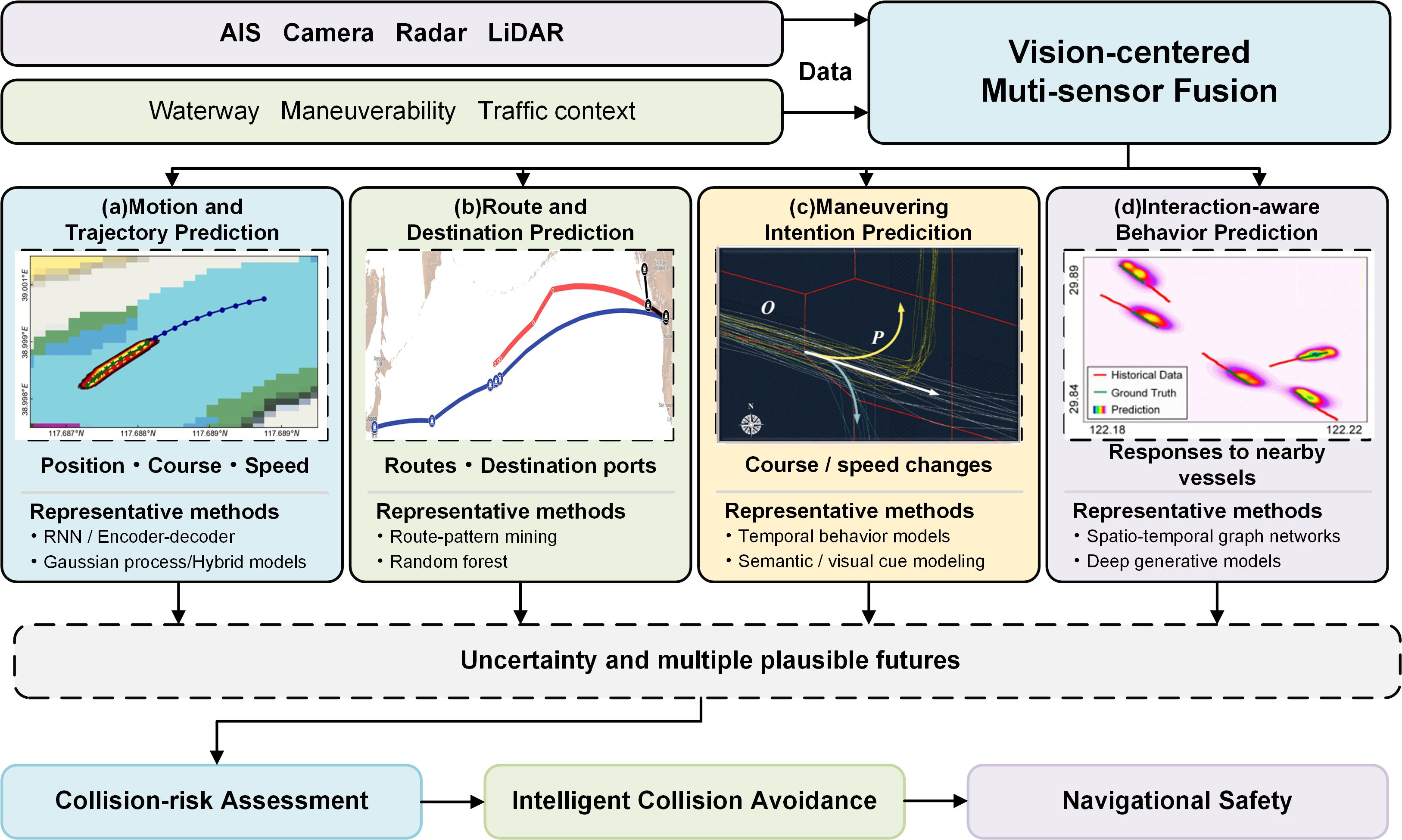}
    \caption{\textbf{Overview of vessel behavior prediction for navigational safety in restricted visibility.} Behavior prediction involves estimating the future movements, trajectories, speeds, and destinations of vessels of interest. It is able to enhance maritime safety, strengthen maritime security, optimize port and traffic management, \textit{etc.}  In current literature, vessel behavior prediction mainly includes motion and trajectory prediction, route and destination prediction, maneuvering intention prediction, and interaction-aware behavior prediction in this work.}
    \label{fig:vessel_behavior_prediction}
\end{figure}

Vessel behavior refers to the movements and operational actions of a vessel, including changes in course and speed, route following, berthing, anchoring, and collision avoidance. These behaviors reflect its navigational goals and responses to surrounding traffic and waterway conditions~\citep{Zhou2019MaritimeTrafficModels}. Predicting such behavior is important for the navigational safety of marine surface vehicles in restricted visibility, where degraded visual observations can limit the timely recognition of developing encounters. Intelligent vision and vision-centered multi-sensor fusion provide complementary information on surrounding vessels, including their locations, motion histories, and contextual cues, which can support forecasts of their future behavior. These forecasts extend situational awareness from the current traffic state to potential future conflicts, supporting anticipatory risk assessment and collision avoidance~\citep{Zhang2022VesselTrajectoryPrediction}. Vessel behavior prediction therefore forms an important link between perception and safe navigation. As summarized in Fig.~\ref{fig:vessel_behavior_prediction}, its principal tasks include motion and trajectory prediction, route and destination prediction, maneuvering intention prediction, and interaction-aware behavior prediction. These complementary tasks address where surrounding vessels may move, which navigational goals they may pursue, and how they may respond to an evolving encounter, with uncertainty considered across the different prediction tasks.

Motion and trajectory prediction estimates a vessel's future positions and, where explicitly modeled, its course and speed over a specified prediction horizon. \citet{Zhang2022VesselTrajectoryPrediction} provides the spatial and temporal basis for traffic monitoring and anticipatory collision-risk assessment. Recurrent neural networks learn temporal dependencies from historical AIS observations~\citep{Capobianco2021RNN}, while recurrent encoder--decoder architectures predict future trajectory sequences together with uncertainty estimates~\citep{Capobianco2022Uncertainty}. Gaussian-process models and hybrid predictors also account for uncertainty in future vessel motion~\citep{Rong2019Uncertainty,Wang2024HybridPredictor}. These approaches extend motion prediction beyond a single extrapolated path by characterizing the range of possible future positions.

Route and destination prediction concerns the broader spatial goals of vessel movement. Route prediction estimates the likely navigational path or sequence of route segments, whereas destination prediction identifies the port or area toward which a vessel is heading. Historical AIS data provide recurring movement patterns that support these tasks~\citep{Tu2017AISSurvey}. Route reconstruction from incomplete trajectories~\citep{Dobrkovic2018RouteReconstruction} and trajectory similarity learning~\citep{Liang2021TrajectorySimilarity} help establish historical route patterns and behavioral references, although neither task alone constitutes future prediction. For destination prediction, AIS-based random forests have been used to infer likely destinations from vessel observations~\citep{Zhang2020DestinationPrediction}. Route and destination information can, in turn, constrain trajectory forecasts by distinguishing movements toward different navigational goals.

Maneuvering intention prediction addresses the actions a vessel is likely to undertake, such as changing course, accelerating, decelerating, stopping, or initiating an avoidance maneuver. It concerns the intended action underlying future movement, rather than only the resulting trajectory. Relevant evidence includes recent motion changes, vessel maneuverability, waterway geometry, and the surrounding encounter situation. Regional ship behavior prediction provides a maritime example of learning behavior from AIS observations~\citep{Murray2021RegionalBehavior}. Related transportation studies have explored semantic information for intention prediction~\citep{TCP} and visual intention cues for trajectory forecasting~\citep{VisualExposesYou}, offering methodological references for incorporating contextual evidence. Their application to maritime navigation nevertheless requires adaptation to vessel dynamics and maritime operating conditions.

Interaction-aware behavior prediction estimates how a vessel's future movement depends on nearby traffic participants. During crossing, overtaking, or head-on encounters, a vessel may alter its course or speed in response to the actions of other vessels. Spatio-temporal graph networks represent dependencies among vessel trajectories~\citep{Liu2022STMGCN}, while interaction-aware deep generative models forecast future motion conditioned on surrounding traffic~\citep{Han2023InteractionAware}. Related forecasting research has developed mechanisms for social interaction modeling~\citep{Alahi2016SocialLSTM,Gupta2018SocialGAN}, vectorized map and agent representations~\citep{Gao2020VectorNet,Liang2020LaneGCN}, and attention-based multi-agent prediction~\citep{Yuan2021AgentFormer,Shi2022MotionTransformer}. These mechanisms offer useful references, but maritime applications must account for sparse and noisy AIS observations, waterway structure, and vessel-specific maneuverability. The International Regulations for Preventing Collisions at Sea (COLREGs) also provide a normative context for encounter behavior, although translating their provisions into computational models remains challenging~\citep{GarciaMaza2022COLREGs,Chang2024COLREGMASS}.

These prediction tasks are complementary: destinations and routes provide navigational goals, maneuvering intentions indicate likely actions, and interaction models describe responses to surrounding vessels. Together, they support trajectory forecasts that account for both individual movement and the evolving traffic situation. Uncertainty should be considered across all these tasks because incomplete observations and alternative behavioral choices can lead to multiple plausible futures. Vessel behavior prediction therefore extends beyond forecasting where a vessel will move to estimating what it is likely to do and how its behavior may evolve within the maritime environment.
\begin{figure}
    \centering
    \includegraphics[width=\linewidth]{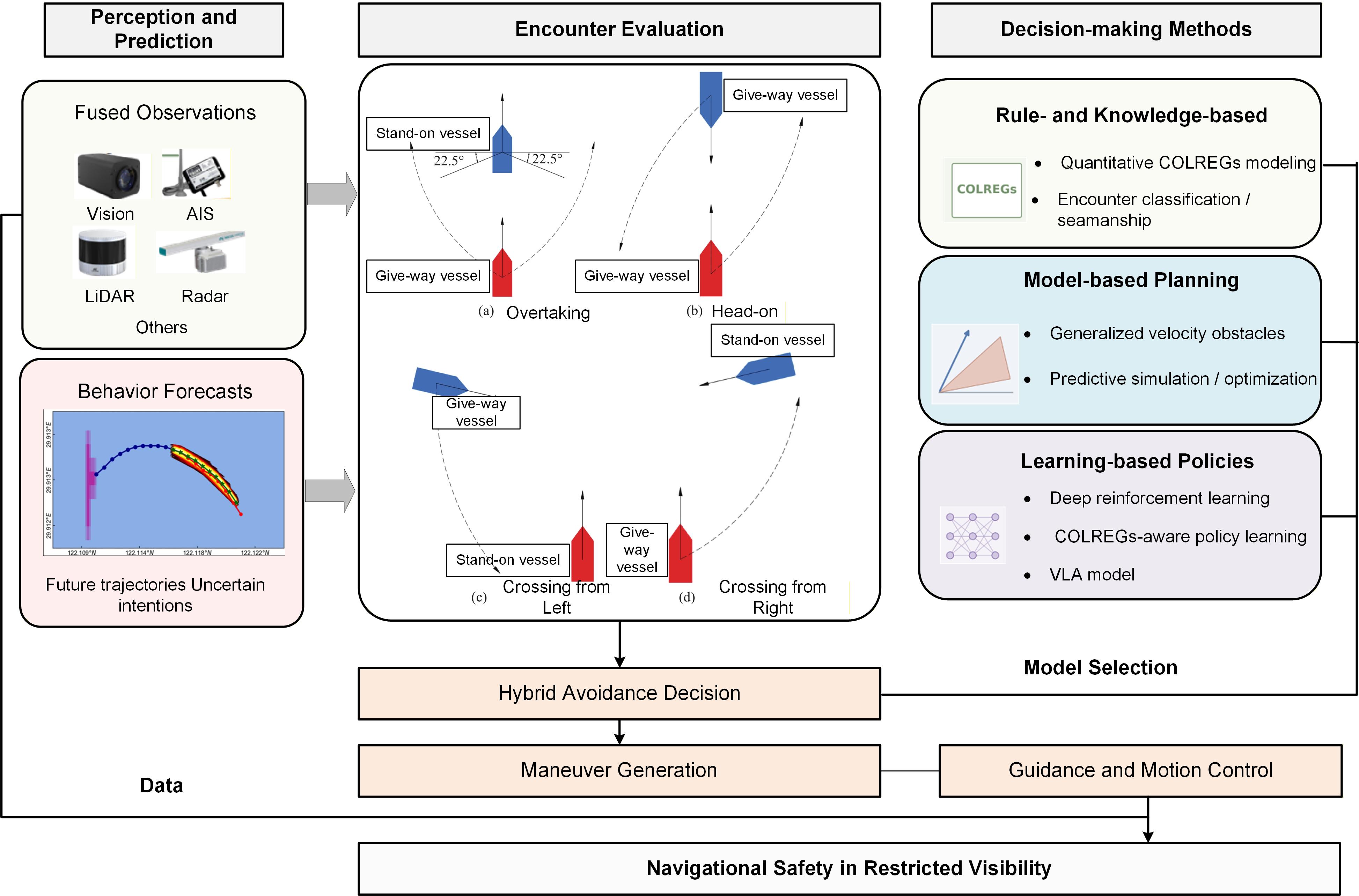}
    \caption{\textbf{Conceptual framework for intelligent collision avoidance in restricted visibility.} Fused observations, behavior forecasts, and operating context support encounter evaluation and collision-risk assessment. Rule- and knowledge-based, model-based, and learning-based methods can be combined to generate avoidance decisions under safety, COLREGs, and maneuverability constraints. Guidance and motion control execute the resulting maneuvers, while updated observations close the feedback loop.}
    \label{fig:intelligent_collision_avoidance}
\end{figure}
\subsection{Intelligent Collision Avoidance}
Intelligent collision avoidance enables marine surface vehicles to assess potential conflicts and select avoidance maneuvers while considering navigational safety, traffic rules, and vessel maneuverability~\citep{Huang2019CollisionAvoidanceSurvey,Zhang2021MASSCollisionAvoidanceSurvey}. Its importance becomes particularly evident in restricted visibility, where degraded visual observations can delay the detection of surrounding vessels and obscure the development of hazardous encounters. Within the scope of this review, intelligent collision avoidance is a downstream application through which intelligent vision and vision-centered multi-sensor fusion contribute to navigational safety. By combining visual information with complementary sensor observations, the perception system can provide the traffic-state estimates needed to assess collision risk and determine appropriate responses. 

Fig.~\ref{fig:intelligent_collision_avoidance} illustrates the closed-loop process connecting upstream perception and behavior prediction with encounter evaluation, avoidance decision-making, and downstream guidance and motion control. Object detection, tracking, and multi-sensor fusion estimate the positions and movements of surrounding vessels, while behavior prediction provides possible future trajectories and associated uncertainties~\citep{Han2023InteractionAware,Capobianco2022Uncertainty,Wang2024HybridPredictor}. These outputs support encounter assessment and avoidance decisions, which specify changes in course or speed, target waypoints, or a planned trajectory. Guidance and control then translate the selected maneuver into executable vessel motion~\citep{Tao2024IntegratedGuidance}. Updated observations allow the system to reassess the encounter as the maneuver proceeds. The effectiveness of this process therefore depends on both the reliability of upstream information and the feasibility of downstream execution.

Collision-risk assessment provides a common foundation for different avoidance methods. Conventional indicators, including distance to closest point of approach (DCPA) and time to closest point of approach (TCPA), characterize the proximity and timing of a predicted encounter. Ship-domain-based approaches further evaluate whether and when another vessel may enter a protected region around the own ship~\citep{Szlapczynski2016DomainRisk}. Probabilistic ship domains represent safety boundaries using data-driven probability distributions~\citep{Zhang2019ProbabilisticDomain}, while environmental risk models incorporate hydrometeorological conditions into ship--ship collision-risk assessment~\citep{Zhang2021HydrometeorologicalRisk}. These developments broaden risk assessment beyond relative geometry, which is relevant when restricted visibility and uncertain observations affect the interpretation of an encounter.

Rule- and knowledge-based methods determine avoidance actions through explicit representations of navigational rules and seamanship. Quantitative modeling of the International Regulations for Preventing Collisions at Sea (COLREGs) supports encounter classification and the selection of rule-consistent maneuvers~\citep{He2017COLREGQuantitative}. This approach makes the reasoning behind an avoidance decision relatively transparent. However, converting contextual rule provisions into computational conditions remains difficult, particularly when several vessels interact or the available observations are incomplete. Critical analyses and recent reviews identify COLREGs interpretation and algorithmic formalization as continuing challenges for autonomous collision avoidance~\citep{GarciaMaza2022COLREGs,Chang2024COLREGMASS}.

Model-based methods generate avoidance maneuvers using explicit representations of vessel motion, encounter geometry, and navigational constraints. Generalized velocity obstacles identify collision-threatening regions in velocity space and support the selection of alternative vessel velocities~\citep{Huang2019GVO}. Simulation-based control selection evaluates candidate behaviors through predictive hazard assessment while considering COLREGs compliance~\citep{Johansen2016SimulationBased}. Multi-ship approaches further address avoidance under simultaneous encounter constraints~\citep{Cho2020COLREG}. These methods provide a structured basis for evaluating candidate actions, but their predictions and decisions depend on the accuracy of vessel-state estimates, motion models, and assumptions about surrounding traffic. Uncertainty from upstream perception and behavior prediction consequently needs to be considered when assessing maneuver safety.

Learning-based methods infer avoidance policies from data or simulated interactions. Deep reinforcement learning has been used to learn avoidance actions for unmanned surface vehicles~\citep{Woo2020DRL}, with subsequent work incorporating COLREGs-related objectives into policy learning~\citep{Xu2020CICA}. More recent studies address the balance between collision avoidance and navigation efficiency in complex environments~\citep{Lou2024BalancedDRL}. These methods can represent nonlinear relationships between traffic observations and vessel actions, although performance depends on the coverage of training scenarios and the transferability of learned policies. For restricted-visibility applications, robustness to degraded or incomplete perception is therefore an important consideration.

Hybrid approaches combine complementary elements of these methods to connect rule interpretation, predictive planning, and executable control. COLREGs-compliant hybrid frameworks integrate avoidance mechanisms for different navigational situations~\citep{Eriksen2020HybridCollisionAvoidance}, while integrated guidance and motion-control methods link anticipatory decisions with vessel execution~\citep{Tao2024IntegratedGuidance}. Recent research also considers uncertain target-ship intentions in avoidance decision-making~\citep{Li2026UncertainIntent}. Collectively, these developments highlight the need to carry uncertainty from perception and behavior prediction into risk assessment and maneuver selection. For navigational safety in restricted visibility, the central requirement is therefore an integrated process in which vision-centered multi-sensor fusion supports reliable traffic understanding, prediction identifies potential conflicts, and avoidance decisions remain feasible under vessel dynamics and applicable navigational rules.
\subsection{Vision-Centered Navigation Systems}
\label{sec:navigationsystems}
%

\begin{figure}
	\centering
	\includegraphics[width=\linewidth]{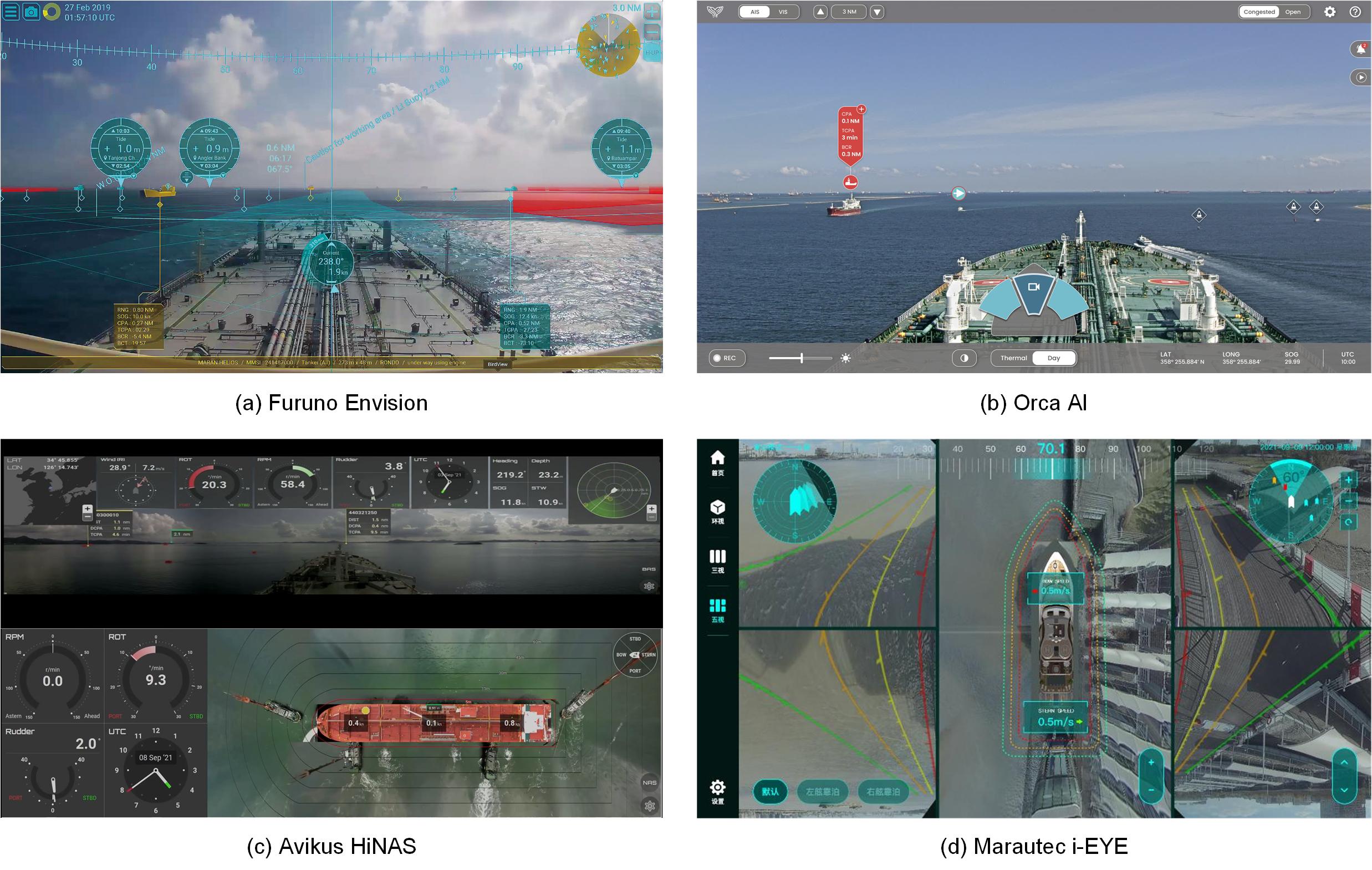}
	\caption{\textbf{Examples of vision-centered navigation systems.} From top-left to bottom-right: (a) Furuno Envision AR-based navigation system; (b) marine situational awareness system developed by Orca AI Ltd.; (c) Hyundai intelligent navigation assistant system (HiNAS) with assisted navigation (top) and berthing/unberthing assistance (bottom); and (d) Marautec i-EYE with several practical capabilities.}
	\label{fig:Figure_systems}
\end{figure}

Vision-centered data fusion has increasingly been incorporated into assisted and autonomous maritime navigation systems, as illustrated in \cref{fig:Figure_systems}. Furuno Envision is an augmented-reality (AR)-based maritime situational-awareness system developed by Furuno Electric Co., Ltd. It combines visible-light cameras, AIS, and radar/chart information to display surrounding vessels and obstacles and support navigation under adverse weather and visibility conditions\footnote{\url{https://www.furuno.com/special/en/envision}}. Orca AI incorporates thermal cameras to complement visual perception under dense haze, nighttime, and reduced visibility\footnote{\url{https://www.orca-ai.io/}}. Satellite communication services such as Starlink can further improve connectivity in remote waters and support maritime operations. 

The Hyundai Intelligent Navigation Assistant System (HiNAS)\footnote{\url{https://www.avikus.ai/}} integrates AI, AR, and multi-sensor data fusion and has been deployed on large vessels, including very large ore carriers (VLOCs), container ships, and liquefied natural gas (LNG) carriers. Its functions include obstacle detection, route following, collision avoidance, and docking assistance. Marautec i-EYE is another representative vision-centered navigation system\footnote{\url{https://www.marautec.com/}}, integrating intelligent vision and data fusion to support target perception and early warning in restricted visibility. 

The Maritime RobotX Challenge has also promoted the development of intelligent navigation systems for USVs. Team Angry-Nerds from the Korea Advanced Institute of Science and Technology (KAIST) developed a USV integrating monocular cameras, LiDAR, and hydrophones and completed five tasks involving navigation and control, underwater search and reporting, autonomous docking, buoy search and observation, and obstacle detection and avoidance~\citep{park2017Development}. \citet{clunie2021development} developed a Robot Operating System (ROS)-based ASV perception system integrating monocular visible-light cameras, LiDAR, and radar for object detection and tracking, semantic segmentation, and localization. 

By integrating wall detection, object detection, path planning, and control, \citet{kim2024field} converted a passenger cruise boat into an ASV and evaluated autonomous navigation in the Pohang Canal. The experiments also highlighted practical challenges, including degraded GPS localization beneath steel bridges and motion disturbances caused by wind and waves. For restricted-visibility navigation, \citet{liu2022intelligent} combined lightweight YOLOX-based visual detection with AIS messages to construct an AR-assisted intelligent navigation system. To improve AR-based maritime interaction, \citet{nordby2024user} introduced a user-interface design concept linking digital navigation information with the physical environment. In addition, \citet{din2023marine} developed the catamaran USV marine$\mathcal{X}$, whose navigation system relies on vision-based target detection and tracking to support operation in complex navigation conditions.

%
%
\section{Open Challenges and Future Directions}\label{sec:challenging}

\begin{figure*}
	\centering
	\includegraphics[width=\linewidth]{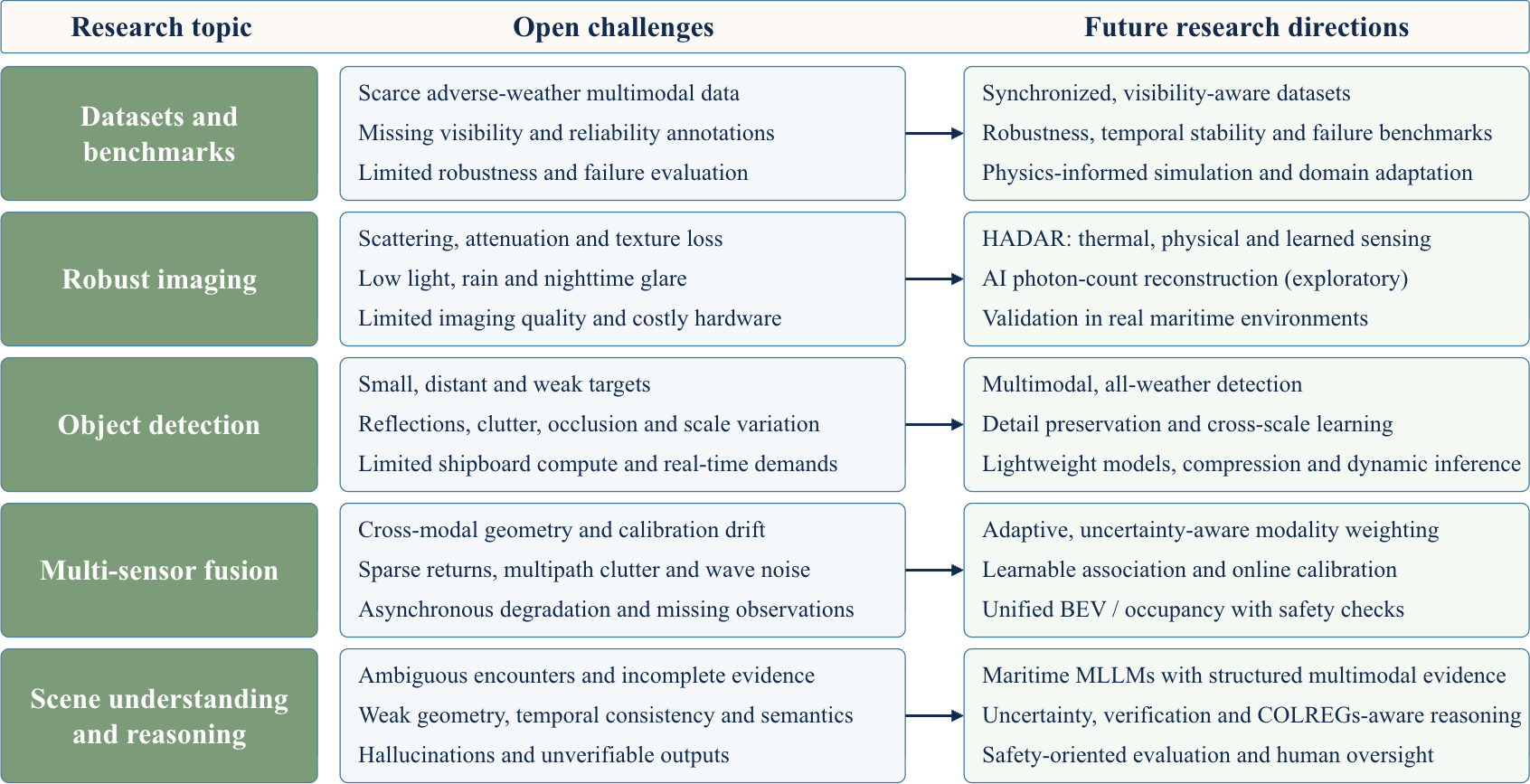}
	\caption{\textbf{Overview of the open challenges and future research directions for navigational safety in restricted visibility.} The five central topics, \textit{i.e.}, datasets and benchmarks, robust imaging in severely restricted visibility, object detection in complex conditions, multi-sensor fusion in dynamic environments, and scene understanding and reasoning, are organized from their key challenges to promising future research directions.}
	\label{fig:challenges_research_directions}
\end{figure*}

Despite the substantial progress reviewed above, guaranteeing navigational safety for marine surface vehicles in restricted visibility still confronts several open challenges and motivates promising research directions. In this section, we organize the discussion around five central topics, as shown in \cref{fig:challenges_research_directions}: datasets and benchmarks, imaging and perception enhancement, object detection, multi-sensor data fusion, and scene understanding and reasoning. For each topic, we first analyze the key challenges and then outline the corresponding future directions.

\subsection{Datasets and Benchmarks for Restricted Visibility}

\paragraph{\textbf{Challenges.}} Dense fog, low illumination, rain, and atmospheric scattering can substantially degrade maritime visual data, resulting in poor visibility, blurred object boundaries, and unstable feature representations~\citep{yao2024waterscenes}. Under such restricted-visibility conditions, optical sensors often suffer from texture loss, low contrast, and disappearance of long-range targets, reducing the reliability of detection, segmentation, and tracking. Maritime environments further introduce dynamic water reflections, irregular target motion, and rapidly changing weather, making perception models sensitive to domain shifts and performance degradation in real-world deployment. Existing datasets, however, remain insufficient for robust perception research in restricted visibility. Most public maritime datasets are collected primarily under clear-weather conditions and rely predominantly on RGB imagery, while relatively few provide adverse-weather scenarios or visibility-aware annotations. Current benchmarks also rarely characterize degradation severity, sensor reliability, or long-range observability in extreme conditions, and the limited availability of large-scale multimodal datasets covering dense fog and other restricted-visibility scenarios therefore remains a major obstacle to reliable all-weather intelligent waterway perception.

\paragraph{\textbf{Future directions.}} Future research should emphasize robust multimodal perception in restricted visibility. Compared with conventional cameras, complementary modalities such as 4D millimeter-wave radar, thermal imaging, and LiDAR can compensate for information loss when optical sensing is severely degraded. Adaptive multimodal fusion, uncertainty-aware perception, and cross-modal feature enhancement are therefore promising directions for improving perception stability in adverse maritime environments. Integrating physics-informed degradation modeling and domain adaptation may further reduce the gap between synthetic and real adverse-weather conditions and improve model generalization.

Another important direction is the construction of comprehensive restricted-visibility maritime datasets and standardized robustness benchmarks. Future datasets should contain synchronized multimodal observations under diverse fog densities, illumination conditions, and navigation scenarios, together with fine-grained annotations of visibility levels, sensor degradation states, and environmental attributes. Beyond conventional accuracy metrics, evaluation protocols should assess robustness, temporal stability, and failure characteristics under severe environmental interference. Automatic annotation, self-supervised learning, and simulation-to-real transfer may further reduce annotation costs and improve dataset scalability, supporting the development of reliable all-weather intelligent waterway transportation.
\begin{figure}
	\centering
	\includegraphics[width=\linewidth]{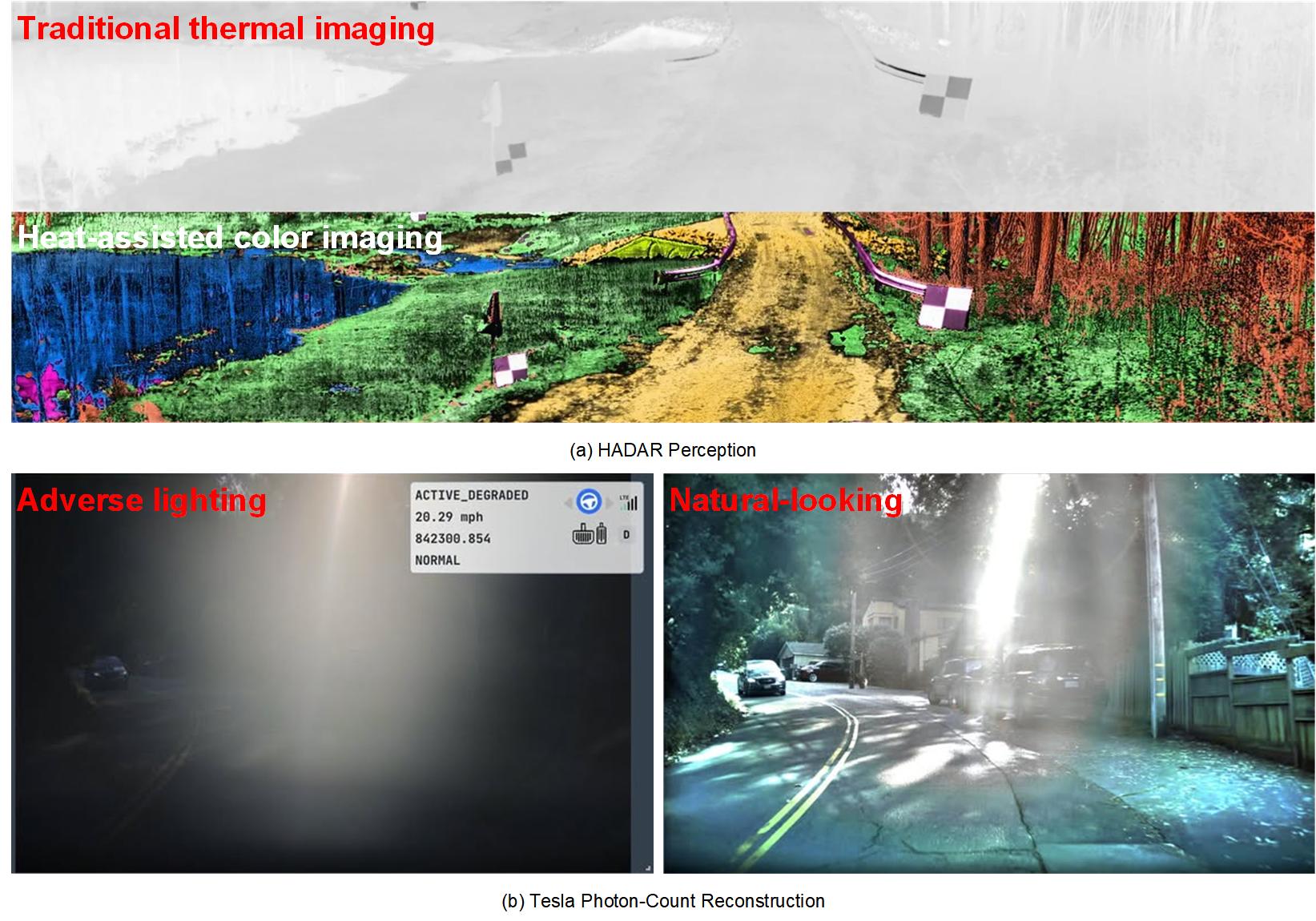}
	\caption{\textbf{Emerging imaging technologies for enhanced visibility.} Illustrations of two advanced imaging technologies, \textit{i.e.}, (a) HADAR and (b) Tesla's AI photon count reconstruction, which can improve MSV perception in dense-haze scenarios.}
	\label{fig:Figure_AdvImaging}
\end{figure}
%
\subsection{Robust Imaging in Severely Restricted Visibility}

\paragraph{\textbf{Challenges.}} Dense haze is common in maritime environments and can contribute to severe navigational risks, economic losses, and casualties. Such conditions induce strong atmospheric scattering and light attenuation that sharply reduce image contrast and visible range, causing distant targets to fade or even vanish, while low illumination, rain, and nighttime glare impose similarly severe yet distinct degradations. Despite extensive efforts in both hardware- and software-based perception, satisfactory imaging under such conditions remains difficult because advanced sensing hardware can be costly, while conventional imaging methods may provide limited visual quality in severely restricted visibility. Consequently, reliable visual perception in such degraded conditions remains a fundamental obstacle, motivating the exploration of emerging imaging technologies.

\paragraph{\textbf{Future directions.}} Emerging imaging technologies originally developed for other domains may provide useful directions for improving MSV situational awareness under adverse weather and imaging conditions. For example, \citet{bao2023heat} presented the heat-assisted detection and ranging (HADAR), which combines thermal imaging, physical modeling, and machine learning for low-visibility perception, shown in \cref{fig:Figure_AdvImaging}(a). In particular, the HADAR extracts physical information from infrared radiation, capable of recovering textural and depth information under low-visibility conditions, providing a potential sensing approach for MSVs operating in dense haze or low-light environments.

Tesla's AI photon-count reconstruction represents another potential direction for recovering scene information when conventional RGB imagery is severely degraded by challenging and dynamic illumination, shown in \cref{fig:Figure_AdvImaging}(b). Its ability to reconstruct visual information under low-light or glare conditions extends naturally to maritime perception in harsh lighting, dense haze, and other restricted-visibility scenarios. Deploying this approach in real maritime environments is a promising direction for robust perception under harsh illumination.

\subsection{Object Detection in Complex Conditions}
\paragraph{\textbf{Challenges.}} Reliable object detection in restricted visibility is challenged by both image degradation and the intrinsic complexity of maritime scenes. Vessels, buoys, and other obstacles exhibit substantial variations in scale, shape, texture, and boundary appearance across distances, viewpoints, and weather conditions. Distant small objects are particularly difficult because they occupy few pixels and often appear as low-contrast, weak-response regions with blurred contours and limited texture. Dense fog, low illumination, rain, and strong water-surface reflections further suppress these cues and can cause confusion with waves, wakes, shoreline structures, and background highlights, leading to missed detections, false alarms, and unstable localization. Partial occlusion, dense target distributions, ship-shore background overlap, and viewpoint changes further reduce detection stability. Existing maritime detectors also face limitations in feature representation, background suppression, and cross-scale detection. Most rely on local appearance cues and conventional multi-scale feature pyramids, which may be insufficient for distant weak targets, low-texture objects, and severe background interference. Deep downsampling can weaken or remove small-object information, whereas shallow features are readily disturbed by water textures, reflection highlights, rain-fog noise, and shoreline clutter, and boundary blurring, dynamic water-surface interference, occlusion, and scale variation can also cause localization errors and unreliable confidence estimates.

\paragraph{\textbf{Future directions.}} All-weather and all-time object detection is important for reliable maritime perception. Because visible-light sensors are vulnerable to adverse maritime conditions, future studies should further integrate infrared, radar, LiDAR, and other sensing modalities to construct multimodal detection frameworks for complex environments. Particular attention should be paid to small, distant, and weak targets in restricted visibility, which often exhibit small spatial extent, weak textures, blurred boundaries, and low target-background contrast. Detectors should therefore improve shallow-detail preservation, high-resolution representation learning, and cross-scale feature interaction. 

Lightweight design and edge deployment are also important for practical MSV applications. Shipboard platforms typically have limited computational resources and strict real-time requirements, whereas many existing detectors remain computationally demanding. Future work could investigate maritime-oriented lightweight architectures, pruning, quantization, knowledge distillation, and dynamic inference to reduce latency and computational complexity while maintaining detection accuracy and robustness.

\subsection{Multi-Sensor Fusion in Dynamic Environments}

\paragraph{\textbf{Challenges.}} Multi-sensor fusion in restricted visibility is further complicated by the pronounced discrepancies among heterogeneous modalities. Navigation radar measures range and velocity in polar coordinates and is commonly represented in a BEV framework, whereas cameras map the environment onto a two-dimensional perspective image, creating a large cross-modal gap in sensing geometry and data representation. Calibration-based alignment is sensitive to drift caused by vessel pitching and long-term vibration, and adaptive matching remains difficult in highly dynamic sea states. For water-surface targets, radar echoes are often sparse and strongly contaminated by multipath clutter, while LiDAR points suffer from wave-induced noise and provide limited appearance information for recognition. Under dense hazy, heavy rain, or nighttime glare, individual sensors may also fail or degrade asynchronously, producing inconsistent, low-confidence, or missing observations. Reasoning over such heterogeneous and partially unreliable evidence while maintaining temporal consistency therefore remains a key challenge for robust fusion.

\paragraph{\textbf{Future directions.}} Future research should develop adaptive and uncertainty-aware fusion frameworks that dynamically reweight modalities according to their instantaneous reliability. Learnable cross-modal association, automatic online calibration, and probabilistic data association can reduce dependence on precise initial extrinsic parameters and accommodate dynamic sensor offsets. Integrating fusion with occupancy modeling and unified BEV representations may further bridge perception and path planning in restricted visibility.

Beyond pairwise fusion, heterogeneous fusion of radar, LiDAR, cameras, infrared, AIS, and other shipboard sensors deserves further exploration, since complementary modalities can compensate for individual failures. Lightweight and edge-deployable fusion architectures are also important for shipboard MSV platforms with limited computational resources. Uncertainty-aware fusion should further be combined with rule-based and probabilistic safety checks to support reliable collision avoidance in highly dynamic maritime environments.
\begin{figure}
	\centering
	\includegraphics[width=\linewidth]{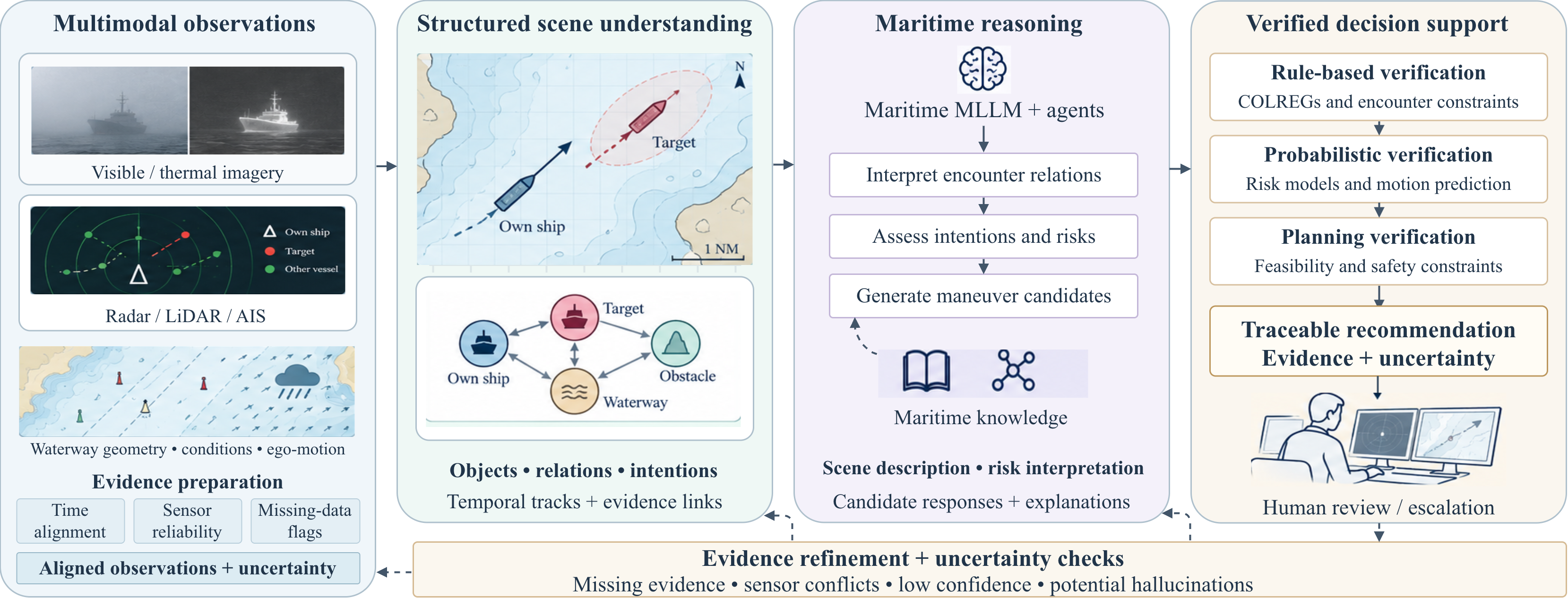}
\captionof{figure}{\textbf{Conceptual framework for maritime scene understanding and reasoning under restricted visibility.} Multimodal observations support scene understanding, maritime reasoning, and verified decision support, with dashed arrows indicating evidence refinement and uncertainty feedback.}
    \label{fig:scene_understanding_reasoning}
\end{figure}
\subsection{Scene Understanding and Reasoning}
\paragraph{\textbf{Challenges.}} Beyond object-level perception, guaranteeing navigational safety further requires scene-level understanding and reasoning, which becomes substantially more difficult in restricted visibility. A safe maneuver depends not only on detecting vessels, buoys, and obstacles, but also on interpreting their spatial relations, encounter geometries, motion intentions, and potential collision risks. Dense fog, low illumination, rain, and atmospheric scattering degrade the very cues on which such reasoning relies: distant targets lose texture and contrast, inter-object spatial relationships become ambiguous, and temporal information is corrupted by motion blur, sensor noise, and intermittent occlusions. Systems must therefore reason over uncertain, incomplete, and sometimes contradictory observations, which greatly aggravates the difficulty of semantic scene comprehension and risk assessment. Existing scene understanding and reasoning methods, especially those built upon vision-language models and multimodal agents, remain insufficient under such conditions. Although recent studies have demonstrated the feasibility of vision-language maneuvering, structured prompt-based risk assessment, and multi-agent waterway question answering~\citep{Kim2025VisionLLMMarine,Kim2025StructuredPrompt,Guan2026WaterVideoQA,Salgado2026USV30}, these methods are mostly evaluated on clear-weather or simulated scenes and may generalize poorly to degraded maritime environments. Current models also exhibit a limited grasp of maritime-specific semantics, weak spatial and geometric reasoning, insufficient temporal consistency, and a tendency to hallucinate when visual evidence is ambiguous, while their outputs rarely quantify uncertainty or provide verifiable evidence.

\paragraph{\textbf{Future directions.}} Multimodal agents provide a potential framework for connecting low-level perception with high-level navigational understanding. For MSVs, navigational safety requires not only detecting ships, buoys, waterlines, and obstacles, but also interpreting encounter relationships, potential risks, and possible responses. Multimodal large language model (MLLM)-powered agents can therefore serve as a semantic reasoning layer above conventional perception modules, translating heterogeneous sensor observations into scene descriptions, risk interpretations, and decision-support information. Early maritime studies have already explored such vision-language reasoning. Vision-LLM-based maneuvering has been investigated in simulated MSV navigation scenarios, demonstrating the feasibility of vision-language reasoning for maneuver selection~\citep{Kim2025VisionLLMMarine}. Structured prompt-based vision-language reasoning further integrates object detection, image captioning, risk assessment, and maneuver recommendation for maritime decision support~\citep{Kim2025StructuredPrompt}. More recently, WaterVideoQA introduced a video question-answering benchmark for waterway scenes and proposed NaviMind, a multi-agent reasoning system comprising routing, captioning, reasoning, grading, and summary agents~\citep{Guan2026WaterVideoQA}. In parallel, USV-3.0 explores cognitive maritime navigation using vision-language models, human feedback, and spatio-temporal navigation memory~\citep{Salgado2026USV30}. These studies indicate an emerging transition from passive perception toward interactive maritime cognition.

\cref{fig:scene_understanding_reasoning} illustrates a conceptual framework connecting multimodal observations, structured scene understanding, maritime reasoning, and verified decision support, with feedback for evidence refinement and uncertainty assessment. Future research on multimodal maritime agents could proceed along several directions. Domain-specific maritime MLLMs could serve as reasoning backbones and be trained or adapted using synchronized multimodal data and structured representations, including visible, thermal, radar, LiDAR, AIS, electronic-chart, weather, and vessel-motion information, and should capture not only object semantics but also encounter geometry, sensor reliability, environmental degradation, COLREGs constraints, and long-horizon temporal evolution. Maritime agents should also be uncertainty-aware and verification-oriented, explicitly characterizing missing observations, inconsistent sensor evidence, low-confidence detections, and potential hallucinations, with safety-critical decisions further checked against rule-based risk models, probabilistic prediction, and formally constrained planning. Future benchmarks should evaluate multimodal agents beyond visual question-answering accuracy, considering factual consistency, temporal stability, COLREGs compliance, risk-ranking correctness, explanation quality, latency, and failure modes under dense haze, heavy rain, nighttime glare, sensor dropout, and challenging traffic encounters. Finally, human-in-the-loop design remains important, with multimodal agents serving as explainable assistants whose interfaces support transparent evidence retrieval, traceable outputs, and conservative escalation when visibility or sensor confidence deteriorates.

\section{Conclusions}\label{sec:conclusions}
This work has provided a comprehensive survey of the techniques for guaranteeing the navigational safety of MSVs in restricted visibility, spanning from intelligent vision to vision-centered multi-sensor data fusion. We first reviewed intelligent vision methods that enhance shipboard perception under adverse weather and illumination, including shipboard video stabilization, visual perception enhancement for hazy, low-light, and multi-scene degradation, object detection and tracking, and navigable area segmentation. We then examined vision-centered multi-sensor data fusion, covering infrared-visible image fusion, radar-vision fusion, LiDAR-vision fusion, AIS-vision fusion, and heterogeneous multi-sensor fusion, and discussed how complementary modalities compensate for the limitations of visible-light cameras in restricted visibility. To ground these techniques in practical scenarios, we further summarized publicly available maritime datasets and analyzed safety-critical applications, including vessel behavior prediction, intelligent collision avoidance, and vision-centered navigation systems.

Despite the substantial progress, guaranteeing navigational safety in restricted visibility still confronts several open challenges, including the scarcity of visibility-aware multimodal benchmarks, robust imaging in severely degraded conditions, small-object detection in complex scenes, adaptive and uncertainty-aware fusion in dynamic environments, and scene understanding and reasoning with vision-language agents. We believe that future research should move from vision-only perception toward uncertainty-aware, rule-aware, and human-in-the-loop navigation. This survey is intended to serve as a structured reference for newcomers and researchers working on the navigational safety of intelligent surface vehicles in IWTS.
\section*{CRediT authorship contribution statement}
\textbf{Ryan Wen Liu}: Writing – review \& editing, Writing – original draft, Supervision, Resources, Funding acquisition.
\textbf{Fengchuan Song}: Writing – review \& editing, Writing – original draft, Validation, Methodology.
\textbf{Tianxiang Wang}: Writing – review \& editing, Writing – original draft, Validation, Methodology.
\textbf{Guobei Peng}: Writing – review \& editing, Writing – original draft, Validation, Methodology.
\textbf{Youngrong Kim}: Writing – original draft, Methodology.
\textbf{Xian Zhong}: Writing – original draft, Methodology.
\textbf{Hooi-Siang Kang}: Writing – original draft, Methodology.
\textbf{Maohan Liang}: Writing – review \& editing, Writing – original draft, Validation, Methodology.
\textbf{Junzhi Yu}: : Writing – review \& editing, Writing – original draft,Supervision.
\textbf{Shanliang Yao}: Writing – review \& editing, Writing – original draft, Funding acquisition.
\section*{Declaration of competing interest}
This manuscript has not been published or presented elsewhere in part or in entirety and is not under consideration by another journal. We have read and understood your journal’s policies, and we believe that neither the manuscript nor the study violates any of these. There are no conflicts of interest to declare.
\section*{Acknowledgments}
This work was supported by the National Natural Science Foundation of China under Grants 52422111, 52271365, and 62271361; the Excellent Youth Foundation of Hubei Scientific Committee under Grant 2024AFA042; the Key Research and Development Program of Hainan Province under Grant ZDYF2026GXJS024; the Key Research and Development Program of Hubei Province under Grant 2024BAB039; the Fundamental Research Funds for the Central Universities under Grant 104972026KFYzxk0029; and Basic Research Program of Jiangsu under Grant BK20251091.

%
\bibliographystyle{cas-model2-names}

\bibliography{cas-refs}



\end{document}